\documentclass{article}

\usepackage{arxiv}

\usepackage[numbers]{natbib}
\usepackage{doi}

\usepackage{float}
\usepackage{graphicx}%
\usepackage{multirow}%
\usepackage{amsmath,amssymb,amsfonts}%
\usepackage{amsthm}%
\usepackage{mathrsfs}%
\usepackage[title]{appendix}%
\usepackage{xcolor}%
\usepackage{textcomp}%
\usepackage{manyfoot}%
\usepackage{booktabs}%
\usepackage{algorithm}%
\usepackage{algorithmicx}%
\usepackage{algpseudocode}%
\usepackage{listings}%

\usepackage{tikz}
\usetikzlibrary{positioning}

\usepackage{comment}
\usepackage{tabularx}
\usepackage{multicol}

\usepackage{enumitem}

\usepackage{cleveref} 
\usepackage{siunitx}

\usepackage{subfig}

\usepackage{tikz}

\definecolor{mixedexpl}{RGB}{103,197,201}
\definecolor{rulebasedexpl}{RGB}{220,176,242}
\definecolor{visualexpl}{RGB}{248,156,116}
\definecolor{numericalexpl}{RGB}{246,207,113}

\definecolor{timeseriesdata}{RGB}{103,197,201}
\definecolor{spectraldata}{RGB}{220,176,242}
\definecolor{pictorialdata}{RGB}{248,156,116}
\definecolor{tabulardata}{RGB}{246,207,113}

\theoremstyle{thmstyleone}%

\theoremstyle{thmstyletwo}%

\theoremstyle{thmstylethree}%

\title{Explainable Artificial Intelligence Techniques for Interpretation of Food Models: a Review}

\author{ \href{https://orcid.org/0009-0006-2494-0349}{\includegraphics[scale=0.06]{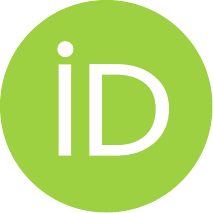}\hspace{1mm}Leonardo Arrighi}\thanks{Corresponding author: \texttt{leonardo.arrighi@phd.units.it}} \\
	Department of Mathematics, Informatics, and Geosciences\\
	University of Trieste, Trieste, Italy\\
	\And
	\href{https://orcid.org/0000-0002-8887-4233}{\includegraphics[scale=0.06]{orcid.pdf}\hspace{1mm}Ingrid Alves de Moraes} \\
	Department of Food Engineering and Technology\\
    University of Campinas,	Brazil \\
    \And
	\href{https://orcid.org/0000-0002-9920-9095}{\includegraphics[scale=0.06]{orcid.pdf}\hspace{1mm}Marco Zullich} \\
	Faculty of Technology Policy Management\\
    University of Technology, Delft, The Netherlands\\
    \And
	\href{https://orcid.org/0009-0000-1505-5774}{\includegraphics[scale=0.06]{orcid.pdf}\hspace{1mm}Michele Simonato}\\
	ASAC s.r.l., Cessalto (TV), Italy\\
    \And
	\href{https://orcid.org/0000-0001-9767-8130}{\includegraphics[scale=0.06]{orcid.pdf}\hspace{1mm}Douglas Fernandes Barbin} \\
	Department of Food Engineering and Technology,\\
    University of Campinas,	Brazil \\
    \And
	\href{https://orcid.org/0000-0002-4988-0702}{\includegraphics[scale=0.06]{orcid.pdf}\hspace{1mm}Sylvio Barbon Junior} \\
	Department of Engineering and Architecture,\\
    University of Trieste, Trieste, Italy\\
}

\hypersetup{
pdftitle={A template for the arxiv style},
pdfsubject={q-bio.NC, q-bio.QM},
pdfauthor={David S.~Hippocampus, Elias D.~Striatum},
pdfkeywords={First keyword, Second keyword, More},
}

\begin{document}
\maketitle

\begin{abstract}
Artificial Intelligence (AI) has become essential for analyzing complex data and solving highly-challenging tasks. 
It is being applied across numerous disciplines beyond computer science, including Food Engineering, where there is a growing demand for accurate and reliable predictions to meet stringent food quality standards.
However, this requires increasingly complex AI models, raising concerns.
In response, eXplainable AI (XAI) has emerged to provide insights into AI decision-making, aiding model interpretation by developers and users. 
Nevertheless, XAI remains underutilized in Food Engineering, limiting model reliability. 
For instance, in food quality control, AI models using spectral imaging can detect contaminants or assess freshness levels, but their opaque decision-making process hinders adoption. XAI techniques such as SHAP (Shapley Additive Explanations) and Grad-CAM (Gradient-weighted Class Activation Mapping) can pinpoint which spectral wavelengths or image regions contribute most to a prediction, enhancing transparency and aiding quality control inspectors in verifying AI-generated assessments.
This survey presents a taxonomy for classifying food quality research using XAI techniques, organized by data types and explanation methods, to guide researchers in choosing suitable approaches.
We also highlight trends, challenges, and opportunities to encourage the adoption of XAI in Food Engineering.
\end{abstract}


\section{Introduction}\label{sec:intro}
Rapid technological advances and the amount of data have made Artificial Intelligence (AI) an essential tool in modern industry and research \citep{kakani_critical_2020,benos_machine_2021,thomas_machine_2022,othman_artificial_2023}.
Food engineering represents a perfect application for AI technology, as food requires in-depth study, processing, and analysis. The large volume of data generated in this field makes AI especially valuable for data analysis. However, the extensive use of AI introduces new questions about its reliability. 

To ensure reliable results, it is essential not only to understand the decision-making process behind the AI model but also to enhance its transparency, auditability, and informativeness \citep{longo2024explainable}. 
Despite this, interpretable AI methods are still not widely adopted in the food sector, highlighting the need for greater focus on model transparency in this field. 
In response to this need, eXplainable AI (XAI) has emerged as an important area of research to increase the reliability of AI model predictions. 
It encompasses techniques aimed at elucidating the behaviour of these models by providing insights into their complex operations.
In food engineering, XAI applications allow accurate identification and validation of critical characteristics in tasks such as contaminant detection, nutritional value estimation, and product authentication, ensuring safety, transparency, and reliability in food quality control. 
This enables greater confidence by model users and customers, identifies potential biases to improve accuracy, and supports the development of new, safer, and better-quality products.

Given the essential role of food in human life, the food industry is keenly interested in applying these techniques to ensure the reliability of AI-driven outcomes \citep{manning_artificial_2022}. 
However, we have identified several gaps in the literature linking XAI with food engineering. 
Firstly, there is a lack of standardization in the terminology and keywords used across various publications, creating challenges for data analysts and food engineers to communicate effectively. 
For instance, terms like ``interpretation'', ``explanation'', and ``comprehension'' are often used interchangeably for similar tasks, particularly when leveraging AI models in food quality research. 
Moreover, there is no comprehensive overview of the current state of the art addressing these differences and providing insights about advantages and drawbacks, which could be important to help non-experts understand the progress and potential of these disciplines in research.
Existing literature already includes several surveys collecting works that compile and organize research on XAI.
However, these contributions address different themes and do not specifically focus on the application of XAI to the food industry.
Some present XAI in a purely theoretical way, without considering concrete use cases \cite{barredo_arrieta_explainable_2020,longo2024explainable,mersha_explainable_2024}.
Additionally, there are surveys on AI applications that do not directly address explainability \cite{survey_dl_food, suvery_gan_food}, as well as narrow studies, such as those focused on fruit \cite{survey_fruit} or plant leaves \cite{survey_plants_pheno}, that are limited to specific contexts.

To address these gaps and respond to the growing need to build reliability in AI outcomes through modern XAI, we propose this survey, which offers a comprehensive overview of XAI applications in food engineering, with explainability explicitly placed at the core of the analysis. 
This survey is primarily intended for researchers, practitioners, and professionals in food science, food industry specialists, and food engineers, who can gain a clear view of the state of the art in XAI applications within the food sector and understand their relevance. 
Additionally, this survey serves as a reference guide for AI developers and policymakers who are involved in designing or evaluating explainable systems related to food safety, quality control, and traceability. 
By presenting various techniques and their applications in different contexts, the survey helps readers easily identify available methods and facilitates the acquisition of new knowledge in the field.

In particular, it offers an overview of current XAI applications in the food industry across major quality-related tasks, such as food safety, nutritional value determination, sensory attributes, authenticity and traceability, sustainability, and healthiness.

\begin{figure}[!tp]
    \centering
    \includegraphics[width=\textwidth]{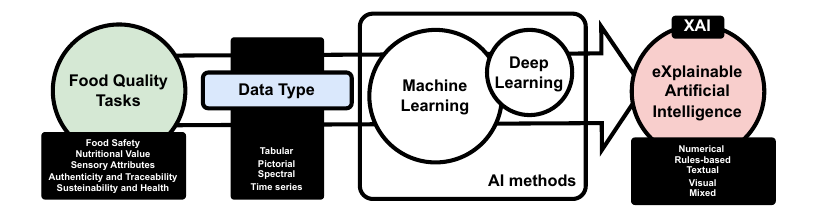}
    \caption{
    Overview scheme, from food quality tasks to XAI techniques.
    XAI is applied as an endpoint of a data processing pipeline that takes into consideration the task, type of data, and the specific AI model employed, e.g., Machine Learning and Deep Learning.
    According to these factors, one or more specific XAI techniques are employed, which produce explanations---tokens of information useful for model developers or users to gain insights into the prediction dynamics.
    Explanations can be produced in different types, each conveying a different facet of the information provided.
    }
    \label{fig:overview}
\end{figure}

To provide a clear reading framework, we categorize applications by data type (tabular, pictorial, spectral, and time series) and forms of explanations generated by the applied XAI methods (numerical, rule-based, textual, visual, and mixed), highlighting their potential for further development, as depicted in \Cref{fig:overview}. 
Each task, data type, and form of explainability is discussed in detail in Sections \ref{sec:foodquality}, \ref{sec:typedata}, and \ref{sec:xaimethods}, respectively.

Additionally, we aim to bridge the gap between XAI and food quality research by proposing a taxonomy and systematically organizing the current landscape of XAI applications in food-related studies. Key contributions of this paper are:
\begin{itemize}
\item to make a comprehensive survey and define a classification system to organize XAI methods applied to food quality;
\item to introduce a taxonomy related to food quality to enhance understanding of the analyzed works;
\item to summarize the XAI techniques used, detailing the types of data and AI methods employed in these studies;
\item to offer an overview of the current state of XAI applications in the food sector, drawing insights from over a hundred papers;
\item to provide comparative insights from the analyzed works, presenting intuitive connections between food quality tasks, data types and XAI methods;
\item to highlight ongoing challenges and to propose potential future research directions in the food industry.
\end{itemize}

Through this survey, we aim to enable scholars and practitioners to identify the most suitable technique based not only on the problem they are facing or the data they possess but also on the intended application and the desired type of explanation.

\subsection{Review methodology}\label{sec:method}
We conducted a \emph{semi-systematic} literature review, combining structured database queries with backward and forward snowballing to improve coverage in a fast-evolving interdisciplinary area.
We queried \emph{Scopus} and \emph{Google Scholar} using combinations of the keywords: ``explainable artificial intelligence'', ``XAI'', ``food'', ``food science'', ``food quality'', ``food control'', and ``agriculture''.
We focused on works published in the last ten years and included studies available up to July 2025.
We applied a targeted eligibility criterion: we retained only studies explicitly addressing \emph{food quality} and explicitly employing at least one \emph{XAI technique}; papers that discussed AI and food but without an explicit XAI method were excluded.
Finally, we expanded the initial corpus through backward screening of reference lists and forward citation chasing, starting from foundational XAI techniques (e.g., LIME, SHAP, CAM/Grad-CAM, PDP, LRP).
\Cref{fig:method_workflow} summarizes the workflow.

\begin{figure}[ht!]
\centering
\begin{tikzpicture}[
  box/.style={draw, rounded corners, align=left, inner sep=6pt, text width=0.9\linewidth},
  arrow/.style={-latex, thick}
]
\node[box] (id) {%
\textbf{Identification}\\
\textbf{Databases:} Scopus; Google Scholar\\
\textbf{Search terms:} ``explainable artificial intelligence'', ``XAI'', ``food'',\\``food science'', ``food quality'', ``food control'', ``agriculture'' (combined)\\
\textbf{Time window:} last 10 years; works available up to July 2025
};

\node[box, below=10pt of id] (screen) {%
\textbf{Screening}\\
Title/abstract screening for relevance to food quality and XAI
};

\node[box, below=10pt of screen] (elig) {%
\textbf{Eligibility}\\
Full-text assessment\\
\textbf{Include:} explicit food quality focus + explicit XAI technique\\
\textbf{Exclude:} food-related AI without explicit XAI method
};

\node[box, below=10pt of elig] (snow) {%
\textbf{Snowballing (iterative)}\\
Backward: references of eligible papers\\
Forward: citation chasing from foundational XAI techniques\\ (LIME, SHAP, CAM/Grad-CAM, PDP, LRP)
};

\node[box, below=10pt of snow] (incl) {%
\textbf{Included studies}\\
Final corpus: $>$200 works surveyed
};

\draw[arrow] (id) -- (screen);
\draw[arrow] (screen) -- (elig);
\draw[arrow] (elig) -- (snow);
\draw[arrow] (snow) -- (incl);
\end{tikzpicture}
\caption{Semi-systematic literature review workflow. 
Structured queries on Scopus and Google Scholar using XAI and food-related keywords were combined with backward and forward snowballing from foundational XAI techniques (e.g., LIME, SHAP, Grad-CAM). Studies were retained only if they explicitly addressed food quality and employed at least one XAI method.}
\label{fig:method_workflow}
\end{figure}

\subsubsection{Methodological limitations}\label{sec:method_limits}
This survey is \emph{semi-systematic} rather than fully systematic: while structured searches were performed, the final corpus also depends on iterative snowballing, which may introduce selection bias toward highly cited works.
Second, restricting the search to Scopus and Google Scholar can omit relevant studies indexed in other databases.
Third, keyword-based retrieval may miss papers that adopt non-standard terminology for explainability or food quality tasks.
Finally, our coverage is time-bounded (up to July 2025) and thus does not include subsequent publications.

\newpage

\section{Explaining Food Quality}\label{sec:foodquality}

The food industry represents a significant sector of the global economy, where monitoring food quality is essential to ensure that food products available on the market are safe, nutritious, and sensorially attractive.
Food quality directly impacts public health, social well-being, and environmental sustainability, influencing responsible production and consumption practices. 
Thus, meeting consumer expectations is fundamental to ensure acceptance, promote brand loyalty and encourage healthy food choices, ultimately supporting commercial success and long-term sustainability \citep{nr131i}.

For a comprehensive analysis of food quality, we propose a taxonomy encompassing five main topics: food safety, nutritional composition, sensory attributes, authenticity and traceability across the supply chain, and sustainability and health within the context of food engineering and nutrition. 
Each of these topics offers a detailed understanding of the elements and challenges that comprise food quality, reflecting consumer needs and expectations \citep{nr131i}.

\textbf{Food Safety}: Food safety involves the assurance that food is free from agents that may pose a health risk.
In addition to implementing rigorous hygiene procedures and sanitary practices to minimize contamination risks, controlling pathogens such as bacteria, viruses, and parasites is fundamental. 
Furthermore, the presence of pesticide residues, heavy metals, and harmful chemical additives must also be strictly controlled. 
Specific regulations limit the concentration of these contaminants in food to ensure consumer safety \citep{nr132i}.

\textbf{Authenticity and Traceability}: The authenticity and traceability of food ensure compliance with legal standards and increase consumer confidence. Identifying and preventing fraudulent practices, such as food adulteration and counterfeiting, is essential to guarantee product authenticity. They not only indicate authenticity but also verify species variety and monitor environmental conditions during cultivation, production, and storage, thereby ensuring food quality and sustainability \citep{nr135i,nr136i,da2024deep,demoraes2024interpretation}.

\textbf{Nutritional Value}: Nutritional value is directly related to food composition and how it impacts human health and well-being. 
Foods rich in vitamins, minerals, proteins, carbohydrates, and healthy fats are essential for the proper functioning of the body and prevent nutritional deficiencies based on their compounds. 
Besides nutritional content, the bioavailability of nutrients is an important quality aspect of food \citep{nr133i}.

\textbf{Sensory Attributes}: The sensory requirements of food are directly perceived by consumers, making them an important means of interaction between products and consumers. Attributes like colour, shape, and taste, along with other appearance attributes, are key indicators of quality and freshness. Sensory standards are crucial for denoting fresh food, which usually has higher nutritional value and consumer acceptability \citep{nr134i}.

\textbf{Sustainability and Health}: Sustainability and health are important for the availability of food with desirable sensory and physicochemical characteristics while also guaranteeing animal welfare, environmental preservation, and consumer health. 
The use of technologies to analyze phenotypic characteristics of plants has promoted more resilient and nutritious crops. 
The implementation of automated processes in food production increases efficiency, reduces waste, and improves food safety \citep{nr137i, nr138i}. 
We differentiate health from nutritional value by defining it more broadly to include disease prevention, immune support, mental health, and the effects of food processing, additives, and potential allergens.

\subsection{Food-industry context and domain-specific challenges}\label{sec:food_industry_context}

The tasks examined in the previous section represent the primary operations routinely addressed by food quality analysis within the food supply chain. However, these tasks present domain-specific challenges that substantially influence both model development and the interpretation of model behavior.
Unlike many controlled laboratory settings, industrial food pipelines are characterized by \emph{biological variability} arising from genetics, physiology, and microbiota, as well as \emph{heterogeneity} in raw materials with respect to size, composition, water activity, and fat-to-protein ratio. Additional sources of variation include \emph{seasonal effects} driven by feed composition, climate, and cultivation conditions, alongside \emph{batch-to-batch fluctuations} attributable to sourcing and production parameters.
Collectively, these factors frequently induce distribution shifts across time, production sites, and suppliers, thereby increasing the risk that models exploit spurious correlations that fail to generalize beyond the training conditions.

In addition, food processing and storage introduce \emph{confounding factors}.
Changes in temperature profiles, packaging type and permeability, oxygen exposure, humidity, and storage duration can alter sensory and physicochemical properties in ways that correlate with the target outcome but are not causally linked to the underlying quality attribute.
For example, a vision model trained for defect detection may inadvertently learn to rely on acquisition artifacts such as lighting conditions, background appearance, or conveyor patterns. Similarly, a spectral model developed for freshness estimation may capture instrument drift, preprocessing choices, or incidentally correlated wavelength regions rather than robust chemical signatures.
These realities motivate the adoption of XAI as a diagnostic layer to support \emph{verification}, \emph{auditing}, and \emph{root-cause analysis} of model predictions under operational conditions.

\paragraph{Illustrative industry scenarios.}
To illustrate the practical relevance of XAI, we present a series of concise scenarios that frequently arise in food industry practice, each highlighting contexts in which XAI can meaningfully support decision-making in realistic operational settings.

\textbf{(S1) Contamination detection through imaging-based inspection. }
In high-throughput inspection settings, such as foreign-body detection, surface contamination assessment, and visible defect identification, XAI can assist inspectors in verifying whether a model focuses on plausible image regions, such as the contaminant area itself, rather than background cues, including packaging elements or lighting reflections \citep{nr46,nr76,nr184}. This capability supports rapid triage, facilitates communication between operators and data scientists, and enables targeted re-acquisition or sensor recalibration when explanations systematically highlight non-causal artifacts as the primary drivers of model predictions \citep{miller2019explanation,nauta_anecdotal_2023,kares2025makes}.

\textbf{(S2) Freshness assessment and shelf-life monitoring across multimodal and spectral data.}
In freshness prediction tasks that rely on spectral measurements or on combined sensor streams integrating temperature and humidity readings with product-level measurements, feature attribution methods can identify which spectral bands or environmental variables are responsible for triggering a quality alarm \citep{nr85,nr105,nr210}. Quality managers can leverage these explanations to assess whether a given prediction aligns with known degradation mechanisms, and to determine whether corrective actions should target process parameters such as cooling-chain control or instead focus on product handling procedures.

\textbf{(S3) Authenticity verification and traceability through tabular and spectral fingerprints. }
In authenticity verification tasks such as geographical origin discrimination and adulteration detection, attribution-based methods can highlight the most influential chemical markers or elemental fingerprints driving a classification decision \citep{nr227,nr231}. These explanations support auditing processes by documenting which indicators contributed to a given outcome, thereby helping laboratories and regulatory compliance teams justify their conclusions and prioritize the selection of confirmatory analyses \citep{act2024eu,panigutti2023role,nr49,nr97}.

\textbf{(S4) Quality grading and process optimization from tabular process variables.}
In grading tasks governed by process and physicochemical variables, explainability tools can reveal population-level trends, such as the influence of temperature or processing time on predicted grade, while also helping operators interpret specific borderline cases. This fosters an actionable understanding of model behavior and supports the definition of standard operating procedures for targeted interventions, as illustrated by the tabular food quality applications surveyed in this work, including quality grading and physicochemical-driven assessments in beverage and dairy production contexts \citep{nr22,nr189,nr194,nr223,nr224}.


\subsection{Data Types}\label{sec:typedata}
With the continuous advancement of technology, food quality analysis has significantly evolved, leveraging the diversity of sensors, methods, and devices to collect data into datasets. 
These datasets encompass various modalities, including tabular data, images or pictorial data, spectral data, and time series data, each offering distinct advantages for analysts in evaluating crucial aspects of food quality. 
The complexity and volume of these data have necessitated AI to process large datasets automatically and identify complex patterns, extracting the maximum useful information from these diverse data.

\textbf{Pictorial data}: Pictorial data allow for clear and intuitive visualization of information, facilitating the communication and understanding of complex data. 
They enable the identification of small defects or imperfections in food, such as stains or deformities. 
Additionally, the images are the results of several non-destructive techniques that support sustainable analysis and monitoring without the need for chemical reagents required in other conversion techniques. 
Pictorial data include \emph{hyperspectral imaging} (HSI), \emph{X-ray imaging}, and \emph{multispectral imaging}, all of which are widely applied in the food quality sector.

\textbf{Tabular data}: Tabular data allow for systematic and clear organization of information, which can simplify statistical analyses and data management. However, complexity can arise from integrating interrelated variables. By using AI algorithms, it is possible to explore these datasets to identify non-obvious correlations and interactions between variables, enabling advanced predictive analyses. 

\textbf{Spectral data}: Spectral data allow for detailed and precise analysis of chemical interactions through the analysis of electromagnetic radiation emitted, reflected, or absorbed at different wavelengths. This makes spectral data a highly accurate tool for detecting small changes in the composition of the analyzed food, providing insights that conventional methods may not reveal. Like pictorial data, spectral data are obtained through ``green'', non-destructive techniques. 
Methods such as \emph{near-infrared} (NIR) and \emph{Raman} spectroscopy, along with \emph{proton nuclear magnetic resonance} ($^1$H NMR), offer high precision comparable to imaging techniques but at a lower computation cost. 

\textbf{Time series data}: Time series data enable continuous and dynamic monitoring of various factors over time. These data capture temporal variations in critical parameters, providing detailed insights into trends and anomalies that may arise at different stages of the production and distribution chain. Additionally, environmental sensors use sequential measurements to establish reference parameters over time.


\subsection{AI Methods}\label{sec:aimethods}
With access to a wide array of pre-built libraries and proven techniques, researchers can adapt various AI methods to address their specific challenges. 
Furthermore, as access to data expands, data analysts---such as chemometricians---can leverage AI methods and enhanced resources to apply their techniques more effectively.
This flexibility enables them to find more efficient and straightforward solutions tailored to their data and the objectives they aim to achieve.

Among the works analyzed, only a few propose using classic AI algorithms, such as \emph{Fuzzy Logic} \citep{nr114,nr115}.
While these algorithms offer the advantage of transparency due to their reliance on well-defined rules, they also demand a deep understanding of the problem and the precise formulation of logical frameworks.

A significant portion of the analyzed articles focuses on using Machine Learning (ML) methods.
\emph{Linear Regression} (LR) algorithms are commonly employed for their effectiveness, simplicity, and complete transparency \citep{nr15,nr98}. 
Similarly, ensemble methods such as \emph{Random Forest} (RF) \citep{nr8,nr12} and \emph{Extreme Gradient Boosting} (XGBoost) \citep{nr100, nr102} are widely favoured for their robustness to outliers and their ability to capture intricate relationships within the data. Although they are generally straightforward to explain, their complexity increases as the number of base learners grows.
Some studies utilize unsupervised ML techniques, such as \emph{k-Nearest Neighbors} (kNN) \citep{nr106} and \emph{Clustering} \citep{nr1,nr107}, which offer transparent and relatively interpretable decision-making processes. 
\emph{Support Vector Machines} (SVMs) are also commonly used techniques \citep{nr10,nr105}, although their decision-making process is more complex and harder to interpret. 
\emph{Extreme Learning Machine} (ELM) \citep{nr41} is also noted for its fast learning speed, though it can be challenging to interpret.

Most of the analyzed works leverage Deep Learning (DL) techniques due to their ability to learn complex patterns and extract valuable information from highly intricate data, such as images.
\emph{Neural Networks} (NNs), among the most widely used models, can achieve outstanding performance even on highly complex problems; however, this comes at the expense of being extremely difficult to interpret.
\emph{Convolutional Neural Networks} (CNNs) are the most widely used method for image analysis because of their effectiveness in generalizing and extracting meaningful features.
By modifying their architecture---such as internal layers or final classifiers—--more specialized networks can be developed, such as \emph{VGG} or \emph{ResNet} for greater robustness \citep{nr2}, \emph{MobileNet} or \emph{EfficienNet} for lightweight applications \citep{nr16}, or \emph{You Only Look Once} model (YOLO) for object detection \citep{nr63}. 
Additionally, DL models that generate or synthesize data, such as \emph{Generative Adversarial Networks} (GANs) \citep{nr21}, \emph{Transformers} \citep{nr26}, and \emph{Autoencoders} \citep{nr76}, are also employed. 
These models, when combined with other techniques or used for feature extraction, can significantly enhance performance. However, while their outputs are often understandable, the models themselves remain difficult to interpret.

\section{XAI methods}\label{sec:xaimethods}

As previously mentioned, XAI revolves around the task of \emph{explaining} (parts of) a model to human stakeholders in an effort to improve the reliability or the transparency of an AI system.
Its outputs are termed \emph{explanations}.
This task is usually split in two sub-categories, \emph{explainability} and \emph{interpretability}, depending on the approach adopted.

\textbf{Explainability vs.~Interpretability}: despite often being used interchangeably, these two notions are slightly different in meaning, as explained by \cite{broniatowski2021psychological}.
The author argues that \emph{interpretability} is concerned with understanding the inner workings of a model, while \emph{explainability} is strictly tied to providing post-hoc, approximate insights on a \emph{prediction} operated by the model.
In other words, interpretability can be seen as an intrinsic property of a model, while an explanation has a post-hoc meaning: it is strictly generated on a (non-)interpretable model after the model has been trained.

We provide a glossary table with this distinction in \Cref{tab:interpretability_explainability}.
We additionally point out that, in the context of the present work, we will mostly refrain from using terms such as ``understandability'' or ``comprehensibility'', since they do not appear to be part of a formal XAI taxonomy, and ``trust'' and ``transparency'', since their meaning extends far beyond the scope of XAI.

\begin{table}[t]
    \centering
    \caption{
    Key taxonomy in XAI: interpretability as an intrinsic model property and explainability as a post-hoc analysis tool.
    }
    \begin{tabularx}{\linewidth}{>{\hsize=0.18\hsize}X >{\hsize=0.82\hsize}X}
        \toprule
        \textbf{Interpretability}
        & 
        Intrinsic property of the model: degree over which the predictive logics of the model are \emph{naturally} understandable to a human.
        \newline
        Main \emph{interpretable} models: LR, decision trees, fuzzy logic.
        \\
        \midrule
        \textbf{Explainability}
        & 
        Creation of \emph{explanations} after (\emph{post-hoc}) a model has been trained: they provide (approximate) insights over (a part of) a (non-interpretable) model to make the prediction itself more \emph{interpretable} to a human.
        An overview of the main explainability methods is given in \Cref{sec2:xais}.
        \\
        \bottomrule
    \end{tabularx}
    
    \label{tab:interpretability_explainability}
\end{table}



\textbf{Accuracy vs.~interpretability trade-off}:
Interpretability, as defined above, has often been depicted as being at odds with \emph{accuracy}\footnote{In this specific case, we use the term \emph{accuracy} as a generic stand-in for the performance of the model in solving the task which it was designed to carry out.}, also termed \emph{expressivity} or \emph{flexibility} of the model \citep{james2013introduction}.
Flexibility refers to the range of complicated patterns that the model can learn. LR is often depicted as a very inflexible model because it can only learn simple linear relationships between predictors; hence, its accuracy will be fairly limited on more complex problems, such as those involving pictorial data.
However, the linear relationship is interpretable by human standards: a single parameter of an LR model indicates the additive effect that a perturbation of the corresponding predictor has on the response.
This makes it straightforward to analyze, for instance, the importance of each variable within the model.

\begin{figure}[t]
    \centering
    \includegraphics[width=.625\linewidth]{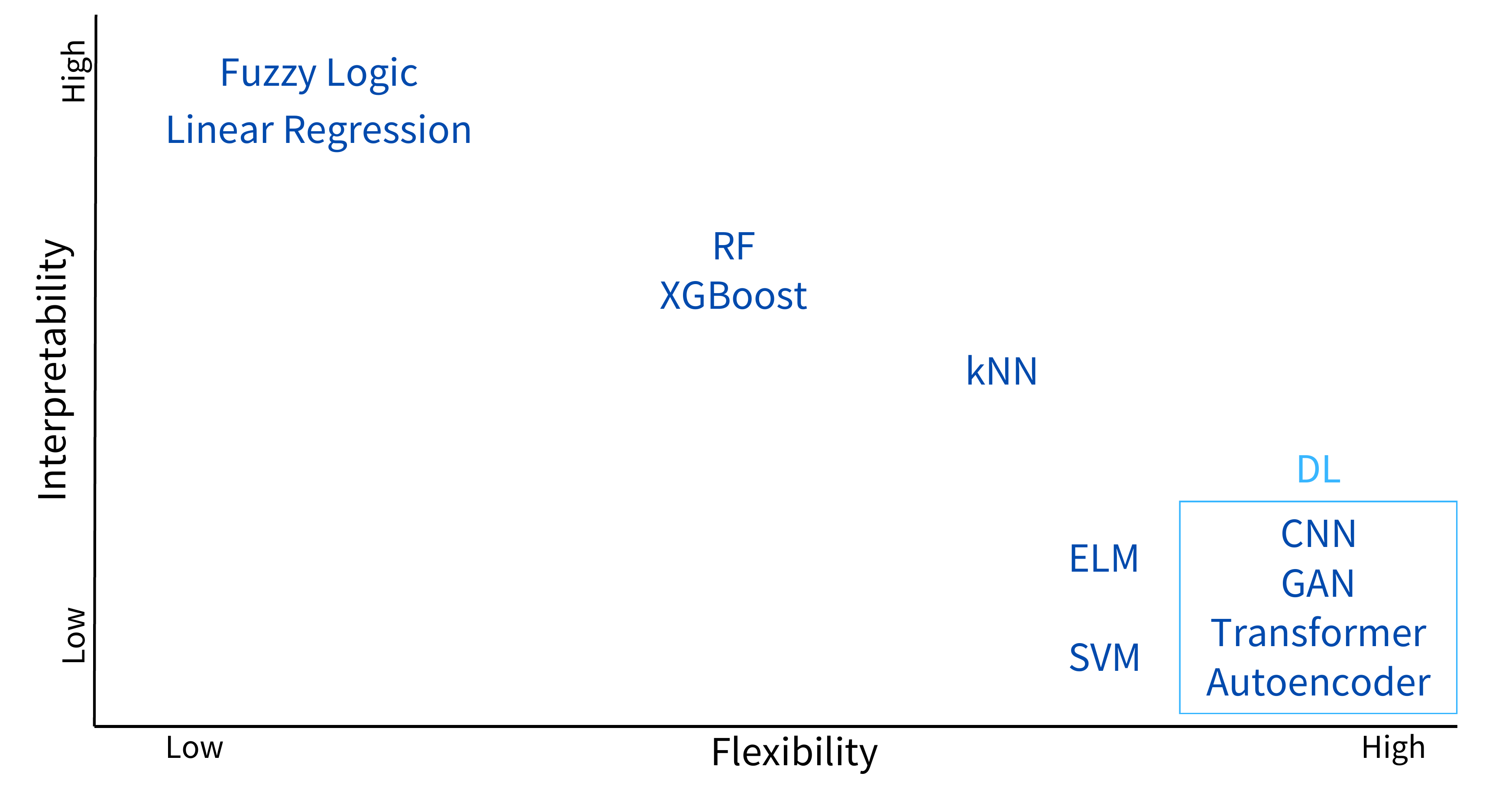}
    \caption{The trade-off between expressivity or flexibility and interpretability of the AI models exploited in the reviewed studies. Expressive models, such as those based on DL, are capable of reaching higher task-level performance but are often difficult to interpret. On the other hand, less complex models, like LR, are inherently interpretable, but often incapable of attaining high task-level performance.}
    \label{fig:interpret-accuracy}
\end{figure}

Conversely, highly expressive models such as Deep NNs are considered complex. 
While they achieve high accuracy on very intricate problems, interpreting the latent rules these models learn to make specific predictions can be quite challenging.
A depiction of this trade-off can be seen in \Cref{fig:interpret-accuracy}, where AI models from the analyzed studies, as described in the \Cref{sec:aimethods}, are positioned accordingly.
Despite the trade-off being renowned in the literature, it is still an approximate rule-of-thumb, which has exceptions, like in the case of vision transformers \citep{dosovitskiy2020image}, which, despite being more expressive than CNNs, are defined as inherently more interpretable due to the ease of visualizing the attention mechanism \citep{shi2024visualization}.

\textbf{Global vs.~local XAI methods}:
Another axis that defines XAI tools is represented by the \emph{scope} of the method.
If the tool delves into properties of the model as a whole, then the scope is said to be \emph{global}; conversely, when the tool investigates the model behavior around one data point, then the scope is said to be \emph{local}.
Concerning the LR example before, the model's coefficients can be thought of as global explanations, since they define a global behaviour of the model irrespective of the specific data point considered.
On the other hand, as an example of local explanation, we can consider \emph{feature attribution} in the context of image classification using CNNs.
For feature attribution, we indicate the action of identifying which variables contribute the most to producing the prediction.
In the case of image classification, it may be of interest to elicit \emph{important} pixels that led a given picture to be classified in a given category; this is an example of a local explanation since we are gaining knowledge of the behavior of the model only on the current image, without trying to infer global properties.
In the case of NNs, it is often hard to identify such global rules for explaining predictions; thus, local explanations are often preferred \citep{wu2020towards}.
Despite being limited in scope, local explanations can be used to extrapolate global information about the models \cite{wu2020towards,setzu2021glocalx}.

\textbf{Model-agnostic vs.~model-specific XAI methods}:
A final property of the XAI tools to be considered is the \emph{specificity} to limited classes of models.
\emph{Model-agnostic} tools are XAI methods that, due to how they are constructed, can be applied to any AI model, while \emph{model-specific} tools are restricted to limited classes of models.

\subsection{Explanation Types}\label{sec:explanations}
\begin{figure}[ht!]
    \centering
    \includegraphics[width=1\linewidth]{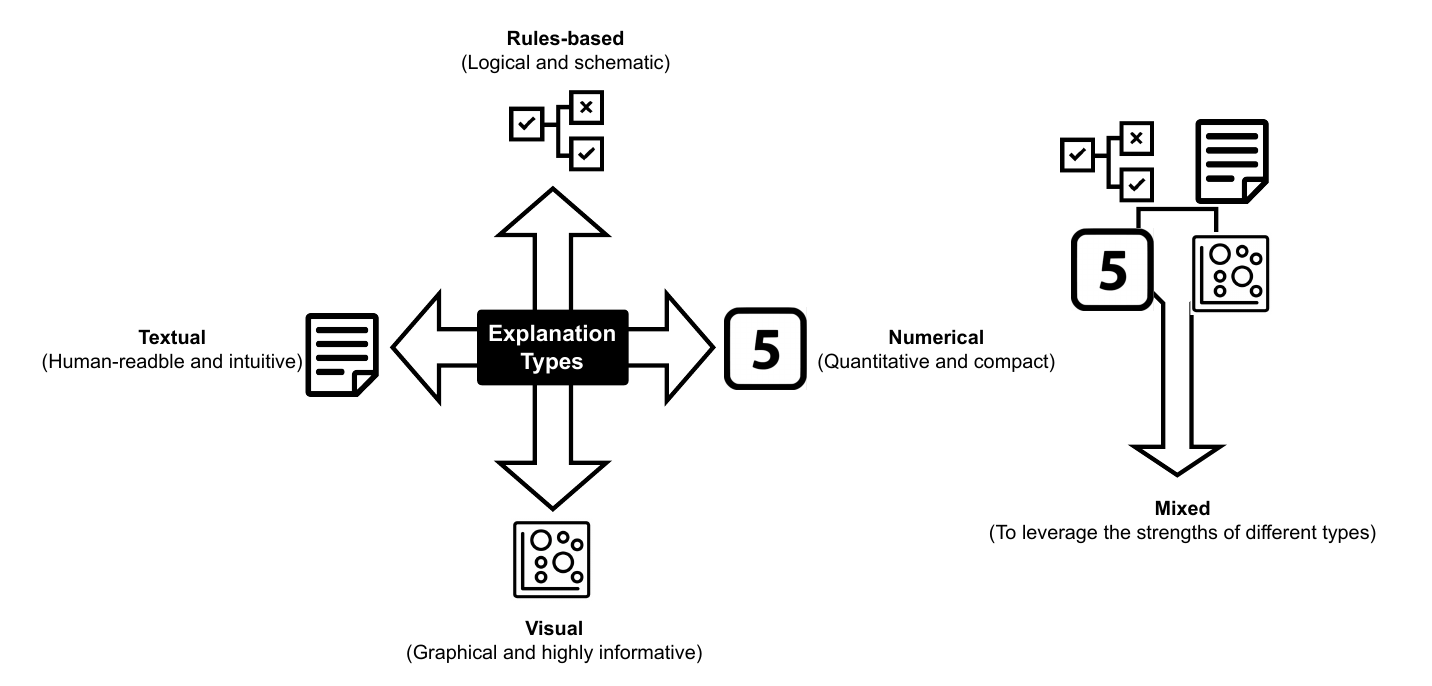}
    \caption{Representation of the types of explanations provided by XAI techniques, along with a summary of their key advantages.}
    \label{fig:datatypes}
\end{figure}

We propose to classify the XAI techniques based on the output format, whether \emph{numerical}, \emph{rule-based}, \emph{textual}\footnote{Since we did not observe any work employing textual explanations, we excluded this type of explanation from the analyses.}, \emph{visual} or \emph{mixed} as is shown in \Cref{fig:datatypes}. 
Different situations may necessitate distinct methods for elucidating the patterns.

\textbf{Numerical explanations}: Numerical explanations are defined as the conveying of information in a compact format using crisp values, vectors of numbers, matrices, or tensors to highlight the input attributes or features of a model that have the largest effect on the prediction of the output. 

\textbf{Visual explanations}: Visual explanations use graphical tools to illustrate information, often through heatmaps, graphs, or other visualizations that highlight specific areas of the data that influence the model's inferential process. 

\textbf{Rule-based explanations}: Rule-based explanations use a schematic, logical format, typically in the form of ``IF$\dots$THEN'' statements with AND/OR operators, to express combinations of input features and their activation values. These rules employ symbolic logic, a formalized system of primitive symbols and their combinations. 


\textbf{Mixed explanations}: Mixed explanations combine multiple formats, such as visual, textual, and numerical explanations, to exploit their strengths and overcome individual weaknesses.

\subsection{Overview of Popular Explainable Artificial Intelligence Techniques}\label{sec2:xais}

In recent years, several explainability techniques have gained importance in response to the growing need for transparent AI models. These methods are generally straightforward to apply and support a range of models and explanation types. 
Together, they constitute a versatile toolkit that can be adapted to diverse cases and operational contexts.
In this Section, we do not detail every technique because of space constraints.
Instead, we just concentrate on the main model-agnostic and model-specific methods, as defined by \cite{vilone_classification_2021}.
We present a concise summary that highlights the key aspects, indicates the explanation type for each method, and provides a general classification based on the taxonomy introduced in the preceding section. 
An overview of these methods is also reported in \Cref{tab2:xais}.

\emph{LIME} \cite{ribeiro2016should} explains a single prediction by training an interpretable \emph{surrogate model}.
This model is trained to mimic the black box model in the neighborhood of the instance to explain. 
It involves perturbing the instance, creating 
$k$ synthetic data points, querying the model on these nearby samples, and weighting them based on their proximity. 
The interpretable model is usually a LR or a small decision tree.
The coefficients of the LR, or the feature importance estimates of the decision tree are used as proxies for the feature importance of the black box model.
These values are numerical explanations, but they can also be plotted (e.g., as a bar plot or a heatmap) to create a visual representation.

\emph{SHAP} \cite{lundberg2017unified} explains a single prediction by attributing to each feature its game-theoretic contribution to the model’s output. 
It compares the prediction for the instance with predictions with one or more features removed, then aggregates each feature’s marginal effect across feature combinations (\emph{coalitions}) to compute its Shapley value.
The feature removal is simulated by replacing the values of the features with a reference baseline (e.g., the average value of each feature).
The explanation is additive, so the sum of feature contributions equals the difference between the model’s expected output and the instance’s prediction. 
While the explanation is numerical, it is typically visualized with dedicated attribution plots.
It can also be visualized as a heatmap, e.g., to explain visual data.

\emph{PDP} \cite{friedman2001greedy} explains model behavior by showing the marginal effect of one or two features on the predicted outcome. 
They are computed by varying the selected feature values over a grid, querying the model at each value while averaging predictions over the distribution of all other features, and assembling the resulting partial dependence function. 
This isolates the average relationship between the chosen features and the prediction, revealing trends such as monotonicity, saturation, or interaction strength. The explanation consists of a line plot for one feature or a surface plot for two features.

\emph{LRP} \cite{bach_pixel-wise_2015}, \emph{CAM} \cite{zhou2016learning}, and its variant \emph{Grad-CAM} \cite{selvaraju2017gradcam} are NN--specific tools.
They explain a single prediction by tracing class evidence back to the input. 
LRP backpropagates the output score through the network with relevance conservation, redistributing relevance at each layer in proportion to local contributions to yield a feature relevance map. 
CAM and Grad-CAM compute a class-specific localization map by weighting the final feature maps with the gradients of the class weights and aggregating them to highlight class-discriminative regions.
CAM and Grad-CAM are originally designed for CNNs for classification trained on image data, although there exist extensions for other architectures, tasks, and data types.
Both methods generate numerical attributions that are typically represented as heatmaps overlaid on the input. 

\begin{table}[tbp!]
\centering
    \caption{
    Overview of the most popular XAI methods classified by scope, specificity, and explanation type. \emph{Efficiency} is an indication of the computational complexity ($\mathcal{O}(\cdot)$) for running the method on a single data point, on which we want to generate an explanation.
    $d$ indicates the number of features, while $k$ indicates the number of additional data points created via perturbation (for LIME).
    }
    \label{tab2:xais}
\begin{tabularx}{\textwidth}{p{0.20\linewidth} p{0.10\linewidth} p{0.18\linewidth} p{0.22\linewidth} p{0.1\linewidth}}
        \toprule
    \textbf{XAI technique} & \textbf{Scope} & \textbf{Specificity} & \textbf{Explanation type} & \textbf{Efficiency} \\
        \midrule
        LIME \cite{ribeiro2016should} & Local & Model-agnostic & Numerical / Mixed & $\mathcal{O}(kd^2)$ \\
        SHAP \cite{lundberg2017unified} & Local & Model-agnostic & Numerical / Mixed & $\mathcal{O}(d2^d)$\\
        PDP \cite{friedman2001greedy} & Global & Model-agnostic & Visual & $\mathcal{O}(d)$\\
        LRP \cite{bach_pixel-wise_2015} & Local & Model-specific & Visual & $\mathcal{O}(1)$ \\
        CAM \cite{zhou2016learning} & Local & Model-specific & Visual & $\mathcal{O}(1)$\\
        Grad-CAM \cite{selvaraju2017gradcam} & Local & Model-specific & Visual & $\mathcal{O}(1)$\\
        \bottomrule
\end{tabularx}
\end{table}

\begin{figure}[t]
    \centering
    \includegraphics[width=13cm]{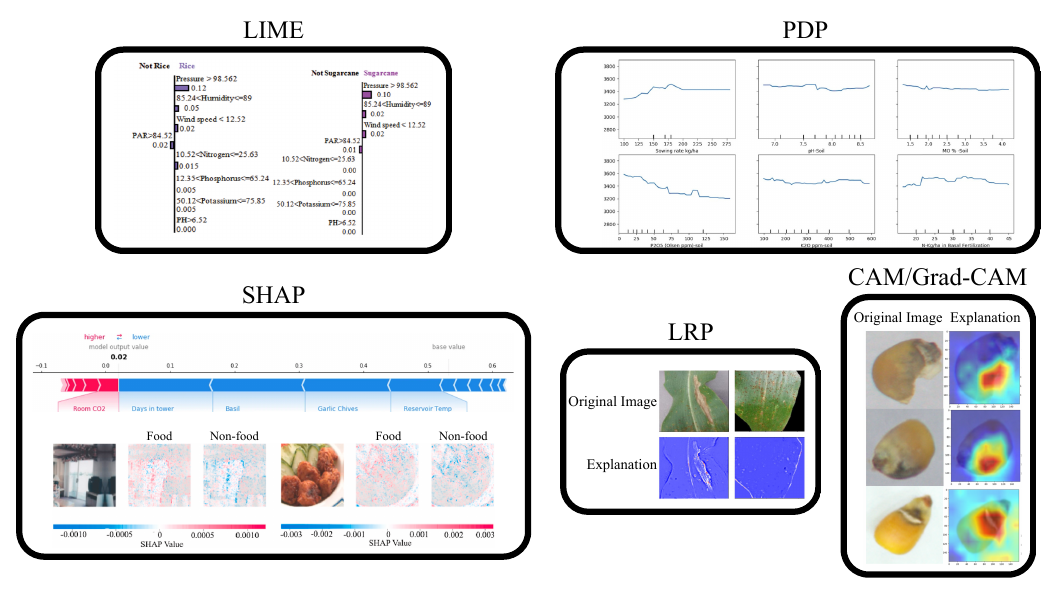}
    \caption{
    Examples of the main techniques introduced in \Cref{sec2:xais}.
    LIME (numerical output shown in vertical bar plots), SHAP (numerical output shown as a \emph{force plot}, and heatmap for image data), PDP (line plots), LRP (heatmap), and CAM/Grad-CAM (heatmap).
    Diagrams are adapted from the following references: LIME, \cite{nr191}; SHAP, \cite{nr36} and \cite{nr24}; PDP, \cite{nr215}; LRP, \cite{nr138}; CAM/Grad-CAM, \cite{nr67}.
    }
    \label{fig:xaitools}
\end{figure}

Overall, \Cref{tab2:xais} highlights how popular model-specific techniques (LRP, CAM, Grad-CAM) have a constant computational complexity with respect to the input dimension; on the other hand, model-agnostic tools, despite their flexibility, all have \emph{at least} a linear complexity.
This provides model-specific methods with a clear efficiency advantage and motivates the widespread adoption of such methods observed in several of the papers surveyed in the present work.
Additionally, in \Cref{fig:xaitools}, we provide an example of visualization for each of the techniques detailed in this Section.
When comparing model-specific tools, most of the works hereby surveyed prefer the usage of CAM-based methods instead of LRP.
It has indeed been observed empirically that human participants tend to find the coarser-grained explanations produced by Grad-CAM more interpretable \cite{kim2022hive,kares2025makes}.
This justifies the wider adoption of Grad-CAM with regards to LRP, despite similar efficiencies.

\section{Explaining Food Safety}\label{sec:summary_foodsafety}
We observe that most of the applications of XAI techniques in the field of Food Safety focus on providing visual explanations, as depicted in \Cref{tab:tbl_foodsafety}. The reason is the frequent use of pictorial data by researchers to study diseases affecting food and insects attacking plants. The images are typically processed using CNNs, with CAM \citep{zhou2016learning} and derived XAI techniques widely applied to explain them.


\begin{table}[ht!]
    \caption{Summary of the works introducing applying XAI for Food Safety topic surveyed in \Cref{sec:summary_foodsafety}, according to their data type and explanation type (labelled as ``Expl.\ type'').}
    \label{tab:tbl_foodsafety}
    \centering    
    {
    \begin{tabularx}{\linewidth}{p{0.3\linewidth} p{0.3\linewidth} p{0.3\linewidth}}
        \textbf{Works} & \textbf{Data type} & \textbf{Expl.\ type} \\
        \midrule
        \citep{nr2,nr11,nr16,nr25,nr26,nr28,nr30,nr32,nr41,nr42,nr47,nr48,nr51,nr52,nr54,nr55,nr57,nr58,nr63,nr65,nr66,nr67,nr71,nr72,nr80,nr81,nr82,nr83,nr86,nr117,nr122,nr123,nr124,nr127,nr129,nr130,nr138,nr143,nr144,nr149,nr150,nr151,nr153,nr156,nr163,nr164,nr166,nr168,nr169,nr170,nr171,nr174,nr175,nr177,nr182,nr187,nr197,nr198,nr199,nr201,nr202,nr205,nr206,nr216,nr218,nr219,nr220,nr221,nr222,nr232,nr233} & \textcolor{pictorialdata}{Pictorial} & \textcolor{visualexpl}{Visual} \\

        \citep{nr31} & \textcolor{spectraldata}{Spectral} & \textcolor{visualexpl}{Visual} \\

        \citep{nr200,nr211,nr228,nr234} & \textcolor{tabulardata}{Tabular} & \textcolor{numericalexpl}{Numerical} \\
        
        \citep{nr6,nr17,nr121,nr176,nr183} & \textcolor{pictorialdata}{Pictorial} & \textcolor{mixedexpl}{Mixed} \\
        
        \citep{nr15,nr195} & \textcolor{tabulardata}{Tabular} & \textcolor{mixedexpl}{Mixed} \\
    \end{tabularx}
    }
\end{table}

\subsection{Visual Explanation for the Food Safety topic}

\paragraph{Pictorial data.}
Numerous studies using pictorial data have focused on detecting plant diseases in staple crops such as maize, rice, and wheat, underscoring the importance of accurate disease identification in the food supply. \cite{nr32} developed a transfer learning methodology enhancing MobileNetV2 with CAM to diagnose plant diseases in maize and rice. \cite{nr66} and \cite{nr58} used CNN models to detect maize and peanut diseases, applying channel attention and pruning techniques to improve feature extraction. 
Similarly, \cite{nr2} applied CNNs with Grad-CAM to distinguish healthy from infected wheat, effectively identifying disease-affected areas, while \cite{nr199} used similar techniques to extend crop disease classification to a resource-constrained setting.
\cite{nr25} introduced C-DenseNet, a modified CNN model, to grade wheat stripe rust severity, validated using Grad-CAM++ \citep{chattopadhyay_grad-cam_2018}. 
\cite{nr42} developed a YOLOv5s-based model with MobileNetV3 and C3Ghost modules to detect \emph{fusarium head blight} (FHB) in wheat, using Grad-CAM.
In addition, to address the complexities of varying disease images, \cite{nr16} improved MobileNetV2 with a Location-wise Soft Attention mechanism and CAM, demonstrating its practical utility for identifying crop diseases under diverse conditions.

In contrast, some studies explored alternative crops and agricultural sectors. \cite{nr82} used high-resolution video data and a CNN-based object detection model to monitor pecan tree health, focusing on \emph{xylella} disease, validated by Grad-CAM to highlight critical canopy features.  \cite{nr30} introduced T-RNet, a Transformer-Embedded ResNet model, for cassava leaf disease detection, demonstrating focus on relevant areas through Grad-CAM visualizations.
\cite{nr232} proposed LeafVisionNet, a new CNN model integrated with SHAP for black gram leaf disease detection, designed to operate in resource-constrained agricultural settings. A similar need to operate in environments with limited data was addressed by \cite{nr233}, who proposed a cost-effective, explainable framework with DenseNet and various XAI techniques, including SmoothGrad, Grad-CAM++, and Score-CAM, for betel leaf disease detection.
\cite{nr206}, \cite{nr218}, and \cite{nr222} aimed to develop a non-destructive and explainable image-based method for identifying adulteration in red chili powder. \cite{nr206} proposed using a GAN to increase the dataset size and a feature selection model to improve the training of AI models, with SHAP and LIME employed to validate the model’s learning. \cite{nr218} and \cite{nr222} used two DenseNet-based models, leveraging Grad-CAM and LIME to explain the results.

Other studies have focused on tomato and potato disease detection, employing diverse models to improve early and accurate identification. 
\cite{nr86} developed an EfficientNet-based model to classify tomato diseases from segmented leaf images, using ScoreCAM \citep{wang_score-cam_2020} for early detection validation. Similarly, \cite{nr71} combined InceptionNet and U-Net, two CNNs, for tomato disease detection and segmentation, validated by ScoreCAM.
\cite{nr143} proposed an ensemble model combining DenseNetMini with Gradient Product optimization and Grad-CAM to enable interpretable disease detection in plant leaves, specifically for tomato and apple plants.
\cite{nr144} introduced DVTXAI, a Deep Vision Transformer model integrated with SHAP, for identifying infections in tomato and potato plants.
\cite{nr149} developed ExE-Net, an Explainable Ensemble Network for potato leaf disease classification, integrating various CNN-based models with XAI techniques---including LIME, SHAP, and Grad-CAM---to enhance the accuracy and interpretability of potato disease identification.

Similarly, \cite{nr153} and \cite{nr219} applied LIME and Grad-CAM to a DenseNet-based model developed for classifying tomato leaf diseases.
Moreover, \cite{nr177} proposed the use of a MobileNetV2 model combined with data augmentation and reweighting techniques for accurate classification of potato leaf diseases on imbalanced datasets, with Grad-CAM used to explain the model’s predictions.
\cite{nr151} introduced a novel saliency-based XAI method using perturbation techniques for object detection, which iteratively refines saliency maps to enhance the interpretability of the applied ResNet model while maintaining high accuracy in classifying potato diseases.
Two other significant works have been proposed by \cite{nr201} and \cite{nr216}, who introduce two new lightweight CNN architectures, XSE-TomatoNet and XLTLDisNet, designed to detect diseases in tomato plants. In both studies, XAI plays a crucial role, utilizing various techniques such as SHAP, LIME, Grad-CAM, and Grad-CAM++. The comparison of these explanations helps validate the developed models.
Finally, \cite{nr187} presented a tomato health monitoring system that integrates YOLOv8 for detection and MobileNetV3 for real-time counting and classification of diseases, with Grad-CAM++ used to explain the model’s predictions.
It is important to mention that CAM-based technology contributed to model verification and highlighted regions with particular texture and color patterns.

A set of research targeted tree and fruit diseases. \cite{nr117} applied CNN models to detect grape diseases using the PlantVillage dataset, validated with Grad-CAM. \cite{nr52} introduced GLD-Det, a MobileNet-based model for detecting guava leaf diseases, confirmed by Grad-CAM for real-time mobile applications. This is an important example of a CAM-based solution in real-time prediction, improving the sample prediction explanation. \cite{nr11} explored the interpretability of CNN models like VGG, GoogLeNet, and ResNet for fruit leaf classification, demonstrating the superior performance of ResNet with Grad-CAM for feature visualization. In another example, \cite{nr72} integrated a novel module into CNN architectures for fine-grained crop disease classification, with Grad-CAM confirming the model’s focus on relevant features.
\cite{nr221} proposed DynLeafNet, a dynamic, computationally efficient, and explainable CNN-based architecture for automatic plant disease recognition, trained on the PlantVillage dataset, with results explained using Grad-CAM++.

In addition to disease detection, several studies shifted attention toward pest detection and pest management in agriculture. \cite{nr124} evaluated Faster-RCNN with a MobileNetV3 backbone for pest identification, validated with Grad-CAM. \cite{nr122} improved pest classification models using genetic algorithms, confirming model efficiency through Grad-CAM visualizations. \cite{nr127} developed ExquisiteNet, a DL model for pest identification validated by Grad-CAM. Additionally, \cite{nr123} used various XAI techniques to provide detailed visual explanations for a lightweight CNN in crop health monitoring. \cite{nr41} utilized LIME with the proposed I-LDD framework, leveraging ELM for fast and robust disease classification on the PlantVillage dataset, accompanied by visual explanations that highlight diseased leaf areas. These two papers show composite solutions, utilizing multiple XAI techniques to provide more insightful explanations.

Some studies extended the use of XAI techniques beyond agriculture. \cite{nr55} applied Grad-CAM to validate an ML method for classifying mercury exposure in fish, supporting food safety beyond agriculture. \cite{nr57} introduced EffiNet-TS, a model based on EfficientNetV2, incorporating an NN to reconstruct images that highlight key symptoms, thereby clarifying the decision-making process. \cite{nr51} proposed a customized EfficientNetB4 model for high-precision classification of chill leaf diseases, validating the model using Grad-CAM. Similarly, \cite{nr48} evaluated the performance of four CNN models, with EfficientNetB4 performing best on a dataset of diseased and healthy plant leaves, confirming the models' focus on critical disease features like rust pustules through Grad-CAM. \cite{nr83} introduced a meta-learning approach for plant disease detection, interpreted with Task Activation Mapping, a CAM-based technique specifically developed for this study. \cite{nr54} developed a convolutional ensemble network using lightweight CNNs like MobileNetV2, validated by Grad-CAM. 
Similarly, \cite{nr220} implemented MobileNetV2, integrating Grad-CAM and LIME for explainability, to develop an end-to-end explainable system for maize disease diagnosis that can be deployed in the field. \cite{nr205} applied SHAP and LIME to explain the results of a CNN developed for early maize leaf disease detection, suitable for mobile deployment.
Furthermore, \cite{nr26} employed a Vision Transformer model for plant disease classification, with Grad-CAM confirming the model’s focus on relevant disease features.

Recent research has significantly advanced the use of DL models for mobile device deployment in agricultural disease detection. \cite{nr63} adapted a YOLOv8n model for real-time wheat ear detection, optimized for mobile devices. \cite{nr81} utilized MobileNetV2 in detecting tomato leaf diseases, emphasizing its suitability for low-end devices in real-world applications. \cite{nr28} proposed the CD-MobileNetV3 model for identifying corn leaf diseases, demonstrating its efficiency for mobile use. Likewise, \cite{nr67} applied the lightweight ShuffleNetV2 model to detect corn seed diseases. These studies validate their models using Grad-CAM for real-time deployment on mobile platforms, highlighting the increasing role of mobile-optimized models in advancing agricultural monitoring and management.

There are numerous applications of AI in the field of fruit quality, particularly those involving the use of XAI to enhance reliability in model predictions.
An important case is presented by \cite{nr174}, who proposed an ensemble learning framework for fruit plant disease detection using multiple DL models, incorporating LIME across all models as an additional tool for result evaluation.
\cite{nr150} proposed a method for classifying various banana diseases---including Cordana, Black Sigatoka, Pestalotiopsis, and Fusarium Wilt---by analyzing leaf images using EfficientNetB0 and employing Grad-CAM to enhance classification accuracy and interpretability.
\cite{nr164} introduced an interpretable AI-based method for localizing mildew symptoms in grapevine using EfficientNetV2S and Grad-CAM.
\cite{nr168} presented LEViT, a Vision Transformer model for tree leaf disease classification, incorporating Grad-CAM to ensure reliable and interpretable results.
An example highlighting the need to apply multiple XAI techniques for reliable results is \cite{nr169}, who developed an AI-based system for date palm classification—capable of identifying diseases and assessing fruit ripeness—using VGG16 in combination with SHAP, LIME, Grad-CAM, and Grad-CAM++.
\cite{nr182} proposed a modified MobileNetV2-based model to improve the classification of cucumber leaf diseases, ensuring result explainability through the integration of LIME. Cucumbers were also analyzed by \cite{nr202}, who proposed using a CNN to examine plant leaf images for disease diagnosis, leveraging Grad-CAM, Grad-CAM++, and Eigen-CAM to explain the results.
\cite{nr170} aimed to improve the explainability of DL models---specifically a ResNet50 model---used in citrus disease detection, by introducing a novel model-agnostic, local explainer for image-based classification called the Multi-objective Genetic Algorithm Explainer (MOGAE).
\cite{nr171} introduced a CNN-based approach for detecting mulberry leaf diseases, utilizing the MobileNetV3Small model and Grad-CAM to align model predictions with expert assessments.
Finally, to enhance crop management by detecting potential diseases, \cite{nr197} used a combination of two versions of InceptionNet and ResNet to classify diseases in orange plants, explaining the results with LIME, Grad-CAM, and Score-CAM, and comparing the obtained outcomes.

More recent applications focus on crops, the primary source of human sustenance, highlighting the growing role of AI and XAI in ensuring food security
\cite{nr80} developed MaizeNet, a CNN framework combining clustering for maize crop image segmentation and classification. Grad-CAM was applied to explain the model, providing severity assessments and crop loss estimation.
\cite{nr65} employed CNNs to quantify rice grain chalkiness caused by high nighttime temperatures, using Grad-CAM to localize affected areas. \cite{nr47} proposed a convolution-based method for rice disease detection, with Grad-CAM highlighting the model's effectiveness even in complex scenarios.
\cite{nr130} proposed a novel DL model that combines DenseNet for feature extraction with an SVM for classifying healthy and diseased sugarcane plants, incorporating LIME to enhance reliability and usability.
\cite{nr138} and \cite{nr175} applied LRP to enhance VGG16 models for identifying crop leaf diseases, aiming to improve performance.
\cite{nr156} developed a deep transfer learning-based framework for diagnosing rice leaf diseases, leveraging various DL models and integrating Grad-CAM to enhance the system's reliability for farmers.
\cite{nr163} incorporated LIME into an EfficientNet-based model to address reliability issues in plant disease classification.
Since Maize Streak Disease poses a serious threat to maize crops, \cite{nr166} introduced a CNN-based framework for its diagnosis, incorporating SHAP and LIME.
Finally, \cite{nr129} proposed a comparative framework integrating Bayesian optimization for hyperparameter tuning across CNN-based models---InceptionNet, MobileNet, ResNet, and RegNet---to diagnose rice plant diseases, leveraging LIME to enhance the interpretability of model behaviour.

\paragraph{Spectral data.}
Considering spectral data, \cite{nr31} developed a method using HSI and DL to assess FHB infection levels in wheat kernels, extracting reflectance spectra and selecting optimal wavelengths. A residual attention CNN classified infection degrees, distinguishing features across infection levels, as confirmed by Grad-CAM. Although spectral data is key for food safety, it does not significantly use visual explanations.

\subsection{Numerical Explanation for the Food Safety topic}

\paragraph{Tabular data.}
Recent studies have introduced numerical explanations to increase the transparency of AI applications for addressing food safety issues. In particular, all the analyzed works used SHAP to identify the most significant features contributing to the models' outputs.
All of these studies relied on tabular data.
\cite{nr234} proposed a scalable and interpretable framework that combined biomonitoring, spatial screening, and CatBoost to assess health risks from airborne organic pollutants affecting honey, demonstrating that honey, together with XAI-enabled ML, can provide actionable and transparent evidence for environmental health management.
\cite{nr228} developed a scalable and interpretable ML framework based on RF that simultaneously predicted daily milk yield and identified health risks in large dairy herds by integrating environmental, nutritional, and physiological data.
\cite{nr211} proposed a hybrid XAI-based feature selection framework that combines Permutation Feature Importance (PFI) and SHAP values to identify the most informative handcrafted features extracted from insect images for classification. PFI and SHAP are used to quantify and rank feature importance, and the final reduced feature set is obtained by taking the intersection of the top-ranked features from both methods, leading to improved classification accuracy.
A work closely related to \cite{nr211} was presented in \cite{nr200}, where the authors extracted handcrafted texture and intensity features from fish images to develop an accurate ML-based system for early fish disease diagnosis.

\subsection{Mixed Explanation for the Food Safety topic}

\paragraph{Pictorial data.}
Several studies have proposed DL methods to address Food Safety issues using pictorial data. However, they applied different XAI techniques than those previously discussed, resulting in distinct explanation types.

Recent advancements in DL have focused on enhancing food safety by employing various XAI techniques to provide insights into model decisions.
\cite{nr6} explored the application of CNNs for plant disease diagnosis, utilizing XAI methods like LIME, Grad-CAM, and SHAP to offer both visual and mixed explanations. 
\cite{nr121} introduced a novel workflow using ResNet18 for pest recognition, which involved segmenting images into meaningful concepts and explaining decisions through weighted directed graphs and concept importance, improving transparency but noting the complexity of explanation generation. 
\cite{nr17} combined DL with semantic web technologies for cassava disease detection, utilizing a Vision Transformer and a semantic model that integrates environmental data. This approach achieved high accuracy and introduced a unique explainability method using knowledge graphs tailored for end users. 
\cite{nr183} proposed using both visual and numerical explanations from LIME to provide localized feature importance, enhancing the transparency of a CNN-based model for classifying rice crop diseases.
\cite{nr176} presented PLD-Det, an improved YOLOv7-based real-time plant leaf disease detection model, incorporating SHAP explanations to enhance transparency and make predictions more interpretable for farmers.

\paragraph{Tabular data.}
Regarding tabular data, \cite{nr15} introduced a novel model for classifying pistachio species by combining feature selection, XAI-based interpretation with LIME, and classification with LR. 
Finally, \cite{nr195} developed a Decision Support System using an ensemble of ML models to predict the risk of wheat powdery mildew in winter wheat crops before visible symptoms appear, thereby supporting proactive disease management in agriculture. The results were subsequently explained using various XAI techniques, including SHAP and LIME.

\section{Explaining Authenticity and Traceability}\label{sec:summary_authenticity}

By addressing the Authenticity and Traceability challenges of the food supply chain, we identified a wider application of XAI techniques. This area emerged as the second most significant food-related task application of XAI.
\Cref{tab:tbl_authenticity} summarizes the works surveyed in this section.


\begin{table}[htbp]
\centering
\caption{Summary of the works introducing applying XAI for the Authenticity and Traceability topic surveyed in \Cref{sec:summary_authenticity}, according to their data type and explanation type (labelled as ``Expl.\ type'').}
    \label{tab:tbl_authenticity}
\begin{tabularx}{\textwidth}{p{0.55\linewidth} p{0.2\linewidth} p{0.2\linewidth}}
    \textbf{Works} & \textbf{Data type} & \textbf{Expl.\ type} \\
    \midrule
    \cite{nr29,nr44,nr64,nr69,nr78,nr90,nr116,nr126,nr131,nr165,nr172} & \textcolor{pictorialdata}{Pictorial} & \textcolor{visualexpl}{Visual} \\
    \cite{nr9,nr137} & \textcolor{tabulardata}{Tabular} & \textcolor{visualexpl}{Visual} \\
    
    \cite{nr49} & \textcolor{spectraldata}{Spectral} & \textcolor{visualexpl}{Visual} \\
    
    \cite{nr10, nr14,nr35,nr36,nr100,nr104,nr109,nr118,nr119,nr135,nr139,nr145,nr146,nr148,nr152,nr154,nr157,nr161,nr173,nr178,nr186,nr189,nr190,nr204,nr215,nr225,nr229} & \textcolor{tabulardata}{Tabular} & \textcolor{numericalexpl}{Numerical} \\

    \cite{nr203,nr207,nr212} & \textcolor{timeseriesdata}{Time series} & \textcolor{numericalexpl}{Numerical} \\
    
    \cite{nr227,nr231} & \textcolor{spectraldata}{Spectral} & \textcolor{numericalexpl}{Numerical} \\
    
    \cite{nr5} & \textcolor{pictorialdata}{Pictorial} & \textcolor{mixedexpl}{Mixed} \\
    
    \cite{nr7,nr8,nr12,nr22,nr102,nr140,nr142,nr147,nr158,nr159} & \textcolor{tabulardata}{Tabular} & \textcolor{mixedexpl}{Mixed} \\

    \cite{nr21} & \textcolor{timeseriesdata}{Time series} & \textcolor{mixedexpl}{Mixed} \\

    \cite{nr97} & \textcolor{spectraldata}{Spectral} & \textcolor{mixedexpl}{Mixed} \\
\end{tabularx}
\end{table}

\subsection{Visual Explanation for the Authenticity and Traceability topic}


\paragraph{Pictorial data.}
Recent studies using pictorial data have advanced the variety traceability and authenticity verification of agricultural products. \cite{nr90} developed a CNN model for herb variability identification, using Grad-CAM to highlight relevant herb parts while ignoring background noise. \cite{nr78} focused on maize seed classification with a ResNet model, while \cite{nr116} applied HSI and DL to classify hybrid okra variability seeds. 
More recently, \cite{nr172} proposed the application of various CNN models to classify fungal species, followed by the use of Grad-CAM to interpret the model predictions

Beyond traceability, several studies addressed the identification of damaged and adulterated products. \cite{nr29} used a ResNet18 model to detect cocoa beans with bad fermentation, with Grad-CAM providing interpretability. \cite{nr69} developed a lightweight CNN, the Soybean Network, to classify damaged soybean seeds, enhancing quality inspection through Grad-CAM visualizations. Meanwhile, \cite{nr64} introduced CondimentNet, an optimized ResNet18 model, leveraging Grad-CAM to detect adulteration in various condiments.

\cite{nr44} and \cite{nr126} emphasized improving agricultural and food production processes for quality and sustainability. \cite{nr44} developed BraeNet, a modified ResNet classifier using 2D and 3D X-ray imaging to detect internal browning in Braeburn apples, demonstrating the practical application of radiography in inline quality sorting. Similarly, \cite{nr126} explored food supply chain optimization, covering plant growth prediction, energy-efficient refrigeration, and expiry date recognition, reinforcing the role of process improvements in maintaining food quality and safety.

\cite{nr165} and \cite{nr131} proposed the use of Unmanned Aerial Vehicle (UAV) aerial imagery as the primary pictorial data source for two similar AI-based applications.
\cite{nr165} explores an interpretable AI-based approach for identifying and mapping weeds and crops using UAV imagery, applying U-Net for segmentation to filter noise and extract key regions, followed by ViT for classification. XAI techniques such as LRP and Pixel Density Analysis are employed in the classification process to enhance transparency.
\cite{nr131} investigated optimal input image conditions for rice yield prediction using CNN models applied to UAV aerial images captured after the mid-ripening stage, assessing the results with XAI techniques such as Gradient-Based Feature Importance Analysis.

\paragraph{Tabular data.}
Using tabular data, \cite{nr9} highlighted the importance of accurate crop yield forecasting in addressing food quality challenges arising from climate change, population growth, soil erosion, and decreasing water resources. The regression model achieved good performance with activation maps to visualize and analyze the features driving the yield predictions, demonstrating that the length of the growing season and conditions such as temperature and sunlight were critical factors.
Similarly, \cite{nr137} presented an ML framework for agricultural drought prediction in the Tapieh Mountains, China, including SHAP analysis to visually highlight the most influential meteorological factors contributing to drought severity.

\paragraph{Spectral data.}
\cite{nr49} proposed using spectral data to address a traceability problem by developing a rapid, non-destructive method for identifying counterfeited beef adulterated with colourants and curing agents. Applying Grad-CAM to spectral data improved the method by generating visual explanations that highlighted key wavelengths influencing the model's decisions.

\subsection{Numerical Explanation for the Authenticity and Traceability topic}

\paragraph{Tabular data.}
Several studies have applied advanced ML techniques using tabular data for crop yield prediction, integrating multiple data sources and employing SHAP for interpretability. \cite{nr36} demonstrated the effectiveness of using SHAP with an AI model for tabular data analysis in aeroponics through data fusion from multiple sensors. 
\cite{nr225} proposed a CNN-based framework for predicting maize yield, focusing on augmenting the often-limited datasets and explaining the results through SHAP and LIME.
Similarly, \cite{nr14} used XGBoost and SVM to analyze factors affecting rice production, explaining model decisions with LIME. \cite{nr100} applied XGBoost for soybean yield prediction, with SHAP highlighting key factors such as near-infrared light and temperature. \cite{nr35} further explored soybean yield estimation, emphasizing the role of the vegetation index using SHAP.

Some studies incorporated satellite and meteorological data for improved predictions. \cite{nr109} utilized Long Short-Term Memory (LSTM), a type of DNN, trained on multisource data, applying Integrated Gradients and SHAP to identify critical factors like enhanced vegetation index and temperature. \cite{nr104} examined the impact of extreme weather on crop yields, revealing sensitivity differences among crops and regions.

Soil water content has also been a focus of ML models in agricultural management. \cite{nr119} introduced TPE-CatBoost, incorporating soil moisture and environmental factors, with SHAP demonstrating model sensitivity to environmental changes. \cite{nr118} used TPE-GBDT to map soil water content across the Yellow River Delta, identifying key variables such as soil texture and vegetation. \cite{nr10} applied SVMs and SHAP to highlight essential factors in digital soil mapping, reinforcing the integration of terrain and geological data for effective agricultural management.
The idea of selecting the most suitable soil has also been explored by \cite{nr186}, who aimed to improve crop quality by classifying different soil types using an ML model, and applied SHAP to highlight the most important features influencing the model's decisions.
Similarly, \cite{nr152} presented an RF model for predicting soil fertility, using SHAP to highlight various physicochemical soil properties that determine fertility levels.
\cite{nr229} proposed an IoT framework that collected soil and weather data to provide real-time crop recommendations, using RF and a NN as predictive models and explaining the results with LIME.
Finally, \cite{nr215} used XGBoost to analyze soil data and predict wheat varieties in Morocco, once again explaining the results with SHAP and PDP.

There is also a substantial body of work focused on the traceability and analysis of environmental conditions to enhance the production and quality of crops such as rice, wheat, and maize, leveraging SHAP or LIME to identify the most influential features utilized by the models in performing the given tasks.
\cite{nr135} proposed an ML model for crop prediction, integrating Genetic Algorithms for hyperparameter optimization and RF for classification, while applying XAI techniques such as LIME and SHAP to enhance classifier interpretability---ultimately supporting farmers in optimizing agricultural planning, reducing crop losses, and improving productivity.
\cite{nr139} presented ML models for crop classification and yield prediction, leveraging XAI techniques such as LIME and Feature Importance to enhance model interpretability. 
Similarly, \cite{nr178} aimed to provide accurate crop yield predictions by using generative algorithms to optimize a DNN, and employed LIME to explain the model’s outputs.
\cite{nr190} proposed a method for selecting optimal crops based on environmental and soil conditions, utilizing Radial Basis Functions and SHAP.
\cite{nr146} introduced XAI-CROP, an ML-based crop recommendation system improved by including LIME to explain predictions, designed to assist farmers in selecting optimal crops by analyzing soil characteristics, historical crop performance, and weather patterns.
A similar tool was developed by \cite{nr145}, who employed various ML models to recommend optimal crops for specific regions, analyzing the results using LIME and SHAP.
\cite{nr161} aimed to enhance the interpretability of AI-driven crop yield predictions by integrating saliency maps and SHAP analysis into KNN models.
\cite{nr157} leveraged an XGBoost model combined with SHAP values to map and understand the influence of weather and soil variables on wheat yield in Eastern Australia.
\cite{nr204} also introduced a tool for predicting subsurface nutrient levels, specifically nitrogen, potassium, and phosphorus, in cabbage cultivation by using plant growth characteristics to train an NN and explaining the obtained results with SHAP and LIME.
\cite{nr173} introduced ML-based regression methods along with XAI techniques---SHAP and LIME---to predict crop yields and assess the impact of climate change on agriculture.

Finally, several studies propose applications similar to those previously discussed, but adapted to different food products.
In particular, \cite{nr154} applied ML models---specifically tree-based ensemble methods---and LIME to classify blackcurrant powders based on image texture features.
\cite{nr189} proposed using ML-based models, such as RF and SHAP, to enhance coffee quality assessment---traditionally reliant on subjective evaluation---by contributing to the standardization of coffee grading.
\cite{nr148} examined the integration of ANN and XAI techniques, such as Feature Importance, to enhance quality control strategies in the agri-food industry, with a specific focus on milk quality classification.

\paragraph{Time series data.}
Some studies use time series as the data type for training AI models. In these cases, all the works employ explanations based on feature importance, calculated by permuting the model’s learned features and assessing how these variations affect the model’s output.
\cite{nr203} developed a method based on a Graph CNN model to predict grain temperatures across different ecological zones, grain types, and granary structures, supporting early warning and management.
\cite{nr212} embedded mathematical equations into a DL model to control the training process and the features learned from soil data, with the aim of predicting soil moisture.

\paragraph{Spectral data.}
Two studies instead used spectral data to address authenticity and traceability tasks, both employing SHAP to provide numerical explanations of their models. \cite{nr227}, using NIR spectra, built a rapid, non-destructive, and interpretable pipeline to evaluate Astragali Radix quality by origin and antioxidant activity. The pipeline consisted of a classifier to authenticate the root’s origin, followed by an SVM regression model to determine its antioxidant activity. Similarly, \cite{nr231} aimed to determine the origin and grade of green tea from elemental fingerprints, using Inductively Coupled Plasma Atomic Emission Spectrometry to collect data from different tea varieties and classifying them with a DNN.

\subsection{Mixed Explanation for the Authenticity and Traceability topic}

\paragraph{Pictorial data.}
In \cite{nr5}, various XAI techniques were applied to enhance the authenticity verification of honey products using HSI, addressing challenges related to high dimensionality and noise through the use of pictorial data.
By integrating multiple XAI algorithms with CNNs, they developed a wavelength selection method to identify the most informative spectral bands, effectively reducing data dimensionality, particularly in classifying honey by botanical origins.

\paragraph{Tabular data.}
Tabular data was explored by \cite{nr12} and \cite{nr7} applied various XAI techniques to enhance the interpretability of ML models in agricultural analysis. \cite{nr12} developed an RF model to assess the influence of biophysical, bioclimatic, and socioeconomic factors on land use for wheat, maize, and olive groves, with Feature Importance, PDP, and LIME identifying key variables such as drainage density, slope, and soil type. Similarly, \cite{nr7} investigated the effects of no-tillage on maize yield using ML and XAI methods, pinpointing critical biophysical and climatic factors.  
\cite{nr102} and \cite{nr22} demonstrated how XAI techniques, when integrated with ML, provide insights into agricultural expansion and product quality assessment. \cite{nr102} applied XGBoost and SHAP to analyze avocado frontier expansion, visualizing key environmental and accessibility factors. \cite{nr22} used XGBoost with SHAP and PDP to evaluate liquor quality in the Vinho Verde region, identifying key chemical attributes influencing product quality.
\cite{nr8} utilized an RF model with LIME to examine the long-term impact of climate variables and soil properties on crop yields in the Coterminous United States. The study identified critical environmental factors affecting yields, demonstrating the value of XAI for understanding complex agricultural data and supporting climate adaptation strategies for stakeholders.
\cite{nr142} employed XGBoost and SHAP to predict annual palm oil yield in Indonesia by analyzing fifteen agrometeorological variables, including rainfall rates, number of rainy days, and soil properties.
\cite{nr140} proposed a Bayesian ensemble model (BM) to analyze the impact of climate on crop yields, effectively separating climate and technological influences while capturing nonlinear climate effects, resulting in high accuracy and interpretable outcomes.
\cite{nr147} explored the application of XAI techniques---specifically LIME and SHAP---to enhance the transparency and user understanding of ML-based models applied to agricultural tabular data, focusing on two case studies: wheat yield prediction and grape yield prediction for wine production.
\cite{nr158} demonstrated that XAI techniques can enhance transparency in food fraud detection by applying LIME, SHAP, and the What-If Tool \citep{wexler_what-if_2020} to DL models.
Finally, \cite{nr159} proposed the application of various ML-based models, including LR, CatBoost, k-NN, and RF, for automated rice classification in Cammeo and Osmancik rice species. To ensure transparency, SHAP and Individual Conditional Expectation (ICE) plots \citep{goldstein2015peeking} were employed.

\paragraph{Time series data.}
Time series data was also explored; for example, \cite{nr21} introduced DeepFarm, a DL framework for managing and predicting agricultural production under uncertainties such as natural disasters and cyber-attacks. Using DL and causal inference, DeepFarm accurately predicted crop yields across U.S. regions, with precipitation anomalies notably impacting corn yields.

\paragraph{Spectral data.}
Using spectral data, \cite{nr97} investigated $^1$H NMR spectra to determine the geographical origins of Asian red pepper powders, employing ML, SVM, and CNN models with dimensionality reduction techniques. Grad-CAM and SHAP provided insights into the decision-making processes, highlighting metabolite distribution variations as key classification factors. This study demonstrated the potential of these models for broader applications in food authenticity verification.

\section{Explaining Nutritional Value}\label{sec:summary_nutritional_value}
Studies on nutritional property explanations reveal a predominant reliance on visual explanations using pictorial data, with minimal use of rule-based methods and occasional mixed explanation types.
\Cref{tab:tbl_nutritional_value} summarizes the studies surveyed in the present section.


\begin{table}[htbp]
\centering
    \caption{Summary of the works introducing applying XAI for the Nutritional Value topic surveyed in \Cref{sec:summary_nutritional_value}, according to their data type and explanation type (labelled as ``Expl.\ type'').}
    \label{tab:tbl_nutritional_value}
\begin{tabularx}{\textwidth}{p{0.55\linewidth} p{0.2\linewidth} p{0.2\linewidth}}
    \textbf{Works} & \textbf{Data type} & \textbf{Expl.\ type} \\
    \midrule
    \cite{nr3,nr24,nr34,nr40,nr45,nr50,nr61,nr68,nr70,nr74,nr77,nr84,nr88,nr91,nr93,nr94,nr95,nr96,nr128} & \textcolor{pictorialdata}{Pictorial} & \textcolor{visualexpl}{Visual} \\

    \cite{nr18,nr37,nr106,nr107,nr160} & \textcolor{tabulardata}{Tabular} & \textcolor{numericalexpl}{Numerical} \\

    \cite{nr13} & \textcolor{spectraldata}{Spectral} & \textcolor{numericalexpl}{Numerical} \\

    \cite{nr1} & \textcolor{pictorialdata}{Pictorial} & \textcolor{rulebasedexpl}{Rule-based} \\

    \cite{nr132} & \textcolor{tabulardata}{Tabular} & \textcolor{rulebasedexpl}{Rule-based} \\

    \cite{nr98} & \textcolor{tabulardata}{Tabular} & \textcolor{mixedexpl}{Mixed} \\
\end{tabularx}
\end{table}

\subsection{Visual Explanation for the Nutritional Value topic}

\paragraph{Pictorial data.}
Several studies, using pictorial data,  have leveraged DL models and XAI techniques to enhance food classification and nutrient estimation. \cite{nr34} applied a weakly supervised VGG16-based CNN for food image segmentation, using Instance Activation Maps to highlight relevant regions. \cite{nr96} introduced the Wide-Slice Residual Network, incorporating slice convolution blocks for improved nutritional evaluation through Grad-CAM visualizations. \cite{nr91} estimated vegetable mass using CNNs and monocular RGB images, while \cite{nr95} utilized attention mechanisms for classifying unlabeled food images from social media.

Some works focused on user-centric approaches for food recommendation and recognition. \cite{nr94} introduced JDNet, a CNN-based model for mobile food recognition, validated through Instance Activation Maps. \cite{nr93} used Grad-CAM to enhance ingredient recognition in a few-shot learning framework, while \cite{nr128} developed PiNet, a multi-task learning framework improving food recommendation by integrating visual and semantic features.

Optimizing food recognition for edge devices has also been explored. \cite{nr24} developed a MobileNetV3-based system, incorporating a user-centered XAI framework with Grad-CAM++ for dietary assessments. \cite{nr77} proposed a big data-driven approach for nutrient estimation, visualizing critical regions with Grad-CAM. \cite{nr74} applied ResNet34 to predict the mechanical properties of Granny Smith apples, using Grad-CAM saliency mappings to reveal biophysical tissue changes.

\cite{nr84}, \cite{nr3}, and \cite{nr40} contributed to dietary assessment and food image recognition. \cite{nr84} introduced the ChinaFood-100 database, evaluating multiple DL architectures and using Grad-CAM to validate nutrient predictions. \cite{nr3} explored oriental food recognition with VGG16 and InceptionNet, revealing model inconsistencies through LIME and Grad-CAM. \cite{nr40} developed a dietary assessment system combining ELM with a SHAP-guided feature selection strategy.

Beyond classification, some studies integrated advanced DL architectures for food analysis. \cite{nr68}, \cite{nr70}, and \cite{nr61} developed non-destructive evaluation and ingredient prediction models. \cite{nr68} proposed the Swin-Nutrition model, a transformer-based framework validated with Grad-CAM. \cite{nr70} used EfficientNetV1 for allergy prediction and food classification, highlighting critical features with Grad-CAM. \cite{nr61} introduced CACLNet, improving ingredient prediction by addressing class imbalance and background noise through Grad-CAM visualizations.

A multi-modal approach has also been explored to enhance nutrition estimation and food recognition. \cite{nr88}, \cite{nr45}, and \cite{nr50} combined diverse data types and learning techniques. \cite{nr50} improved nutrition estimation using ResNet101, integrating multiscale image and depth data features. \cite{nr45} introduced DPF-Nutrition, a transformer-based approach that generates depth maps for enhanced nutrient estimation. \cite{nr88} developed MVANet, a multi-view attention-based CNN incorporating ingredient and recipe semantics, validated with Grad-CAM for food recognition in healthcare applications.

\subsection{Numerical Explanation for the Nutritional Value topic}

\paragraph{Tabular data.}
Tabular data was discussed in \cite{nr106}, \cite{nr37}, \cite{nr107}, and \cite{nr18}, which apply ML techniques to various food-related challenges. \cite{nr106} used XGBoost to estimate added sugar content in foods, with SHAP enhancing model transparency for consumer awareness in regions without mandatory labeling. \cite{nr37} developed the Flavonoid Astringency Prediction Database, employing ML models like RF to explore the relationship between molecular structures and flavor properties. Similarly, \cite{nr107} applied ML to differentiate pepper spices during storage, using SHAP to identify key organic compounds. \cite{nr18} developed an XGBoost-based model for predicting drug-food interactions using molecular fingerprint similarities, with SHAP providing insights into influential features relevant to clinical applications and dietary planning.
\cite{nr160} proposed a graph-based ML approach to predict the outcomes of formulation trials, aiming to reduce laboratory experiments, material waste, and development time in food design. To enhance interpretability, they applied GNNExplainer \cite{ying_gnnexplainer_2019}, a global explanation method tailored for Graph NNs.

\paragraph{Spectral data.}
Spectral data was explored in \cite{nr13}. The authors employed visible NIR point spectroscopy to estimate sugar content in grape varieties at different maturity stages. Regression ML algorithms and a CNN were applied, with XAI techniques such as Variable Importance in Projection and Gini Importance validating the models and identifying key spectral features.

\subsection{Rule-based Explanation for the Nutritional Value topic}

\paragraph{Pictorial data.}
\cite{nr1} exploited pictorial data to propose a similarity score based on user community preferences, enhancing recommendation quality. The rule-based explainability method assigned each image to an appropriate food diet based on user profiles, supporting personalized dietary recommendations.

\paragraph{Tabular data.}
\cite{nr132} presented a novel no-code methodology for developing predictive models to classify the antioxidant activity of phenolic compounds, leveraging Decision Tree-based algorithms and Conceptual Density Functional Theory (CDFT) descriptors. The resulting models achieved high accuracy and full explainability through explicit, interpretable if–then rules derived from molecular features.

\subsection{Mixed Explanation for the Nutritional Value topic}

\paragraph{Tabular data.}
Tabular data was explored in \cite{nr98}, introducing the Taste Peptide Docking Machine, a computational framework for predicting umami and bitter tastes in peptides. The framework integrates machine learning algorithms with molecular representation schemes, including docking analysis, molecular descriptors, and molecular fingerprints. SHAP and LIME were applied to enhance interpretability, providing insights into key molecular features influencing taste prediction.

\section{Explaining Sensory Characteristics}\label{sec:summary_sensory}

Sensory characteristics are extremely important for quality control, leading to the widespread use of sensors designed to mimic human senses. Among these, spectral devices---commonly used and established in the industry---offer rich information, suggesting potential applications for XAI techniques.
However, most studies focus on pictorial data and visual explanations, with only two works to date addressing spectral data for explainability.
In \Cref{tab:tbl_sensory} we summarize the studies surveyed in the present section.


\begin{table}[htbp]
\centering
    \caption{Summary of the works introducing applying XAI for the Sensory Characteristics topic surveyed in \Cref{sec:summary_sensory}, according to their data type and explanation type (labeled as ``Expl.\ type'').}
    \label{tab:tbl_sensory}
\begin{tabularx}{\textwidth}{p{0.55\linewidth} p{0.2\linewidth} p{0.2\linewidth}}
    \textbf{Works} & \textbf{Data type} & \textbf{Expl.\ type} \\
    \midrule
    \cite{nr43,nr46,nr53,nr59,nr60,nr73,nr75,nr76,nr79,nr181,nr184,nr185,nr179,nr208} & \textcolor{pictorialdata}{Pictorial} & \textcolor{visualexpl}{Visual} \\

    \cite{nr85,nr108} & \textcolor{spectraldata}{Spectral} & \textcolor{visualexpl}{Visual} \\

    \cite{nr19,nr38,nr188,nr194,nr223,nr224} & \textcolor{tabulardata}{Tabular} & \textcolor{numericalexpl}{Numerical} \\

    \cite{nr209} & \textcolor{tabulardata}{Tabular} & \textcolor{rulebasedexpl}{Rule-based} \\
\end{tabularx}
\end{table}

\subsection{Visual Explanation for the Sensory Characteristics topic}

\paragraph{Pictorial data.}
The studies highlight advancements in fruit integrity assessment using DL models and XAI techniques over pictorial data. \cite{nr76} employed X-ray radiography and DL methods, including autoencoders and CNNs, for deep anomaly detection of internal defects such as browning and cavities, with heatmaps enhancing interpretability. \cite{nr59} introduced MBNet, a CNN-based model utilizing sensory data from multiple cameras for pear evaluation. \cite{nr53} applied UNet with synthetic data for internal pear defect segmentation, validated through Grad-CAM heatmaps. \cite{nr79} used DenseNet201 for fruit quality classification, with Grad-CAM confirming its focus on relevant features. \cite{nr43} investigated bruise detection in plums using HSI and CNNs, with Grad-CAM visualizations validating model predictions.

Food freshness assessment has also benefited from DL and HSI. \cite{nr73} developed a VGG16-based model to classify shrimp freshness from smartphone images, using Grad-CAM to confirm inference regions. Similarly, \cite{nr46} employed a colourimetric sensor and RGB images to monitor salmon freshness, with Grad-CAM revealing that the CNN prioritized sensor data over visual texture, emphasizing odor's role in freshness detection.

Beyond fruit, cereal integrity has been explored using XAI methods. \cite{nr75} applied Grad-CAM in an EfficientNet-B3-DAN model to detect rice germ integrity, confirming the model’s focus on relevant features. 
\cite{nr60} addressed crop yield estimation by developing an Inception-ResNet-based regression model for leaf counting, handling occlusions in monocots. Grad-CAM analysis confirmed its focus on leaf tips, validating effectiveness across sorghum and maize datasets.
\cite{nr185} enhanced crop classification by employing a MobileNetV2 model validated with Grad-CAM to assess the visual standard quality of tomatoes, classifying them as \emph{damaged}, \emph{old}, \emph{ripe}, and \emph{unripe}.

The theme of product freshness is also explored in \cite{nr184}, \cite{nr181}, \cite{nr179}.
\cite{nr184} introduced a DL-based model to classify meat freshness into fresh, half-fresh, and spoiled categories, incorporating Grad-CAM++ to support transparent decision-making.
\cite{nr181} presented an InceptionV3 model combined with LIME for efficient and transparent classification of chicken meat freshness, which, when integrated with a robotic arm, enhances automation and food safety in poultry processing.
\cite{nr179} utilized CNN-based models to predict the quality of seabream---categorized as fresh, moderate, or spoiled---based on eye and gill images taken under refrigerated conditions, incorporating LIME and Grad-CAM for model interpretability.

A different study, which employed similar techniques, specifically a ResNet with Grad-CAM to explain the results, was proposed by \cite{nr208}. The authors introduced a framework to monitor the storage stability of beeswax oleogel using computer vision based on DL trained on polarized-light microscopy images of the oleogel.

\paragraph{Spectral data.}
Two studies used XAI techniques to analyze spectral data to address sensory characteristic problems. \cite{nr85} developed a CNN model to classify beef freshness using myoglobin data and reflectance spectra, achieving high F1-scores. Grad-CAM highlighted key wavelength regions, confirming myoglobin’s importance in freshness classification. The method demonstrated robustness against environmental factors, indicating strong industrial potential. Similarly, \cite{nr108} used surface-enhanced Raman spectroscopy and a CNN-based model, the Dual-Branch Wide Kernel Network, to classify bacterial signals.

\subsection{Numerical Explanation for the Sensory Characteristics topic}

\paragraph{Tabular data.}
Several works address Sensory Characteristics topic with the goal of explaining model outputs through numeric explanations, despite their differing approaches and applications based on tabular data.
\cite{nr19} focused on predicting boar taint, an undesirable taste and odor found in the meat of male pigs. Using CatBoost, a tree-based ensemble model, the authors achieved peak performance. SHAP analysis identified key factors correlated with boar taint, including feed type, ventilation system, pharmaceutical treatment, and lairage waiting time.\cite{nr38} developed DL models to classify sweet, bitter, and umami molecules, employing a DNN with molecular descriptors and a Graph NN, achieving similar accuracies. SHAP analysis was applied to interpret DNN predictions, revealing key molecular binding properties.
\cite{nr188} developed an ML-based method using various regression models, including XGBoost and RF, alongside Feature Importance analysis, to predict aroma partitioning in dairy matrices and support food reformulation efforts.
\cite{nr223} and \cite{nr224} proposed two similar applications of an AI framework for modelling and interpreting the rheological behaviour of plant-protein systems under varying processing and thermal conditions. More specifically, they apply regression models for predicting shear stress, viscosity, and temperature-dependent viscoelastic properties of sesame protein isolate gels from processing, thermal, and compositional variables. Moreover, SHAP is applied to explain the outcomes.
\cite{nr194} introduced an ML-based method for predicting dry matter content in Emmental cheese, again using SHAP to explain the obtained results.

\subsection{Rule-based Explanation for the Sensory Characteristics topic}

\paragraph{Tabular data.}
There is a single example of applying XAI to Sensory Characteristic tasks, and it is based on tabular data. \cite{nr209} developed an intrinsically interpretable model to predict crab weight using easily measurable physical features. The proposed model was based on genetic programming, a class of evolutionary algorithms that automatically evolve computer programs (often represented as trees) to solve a task using operations analogous to biological evolution, such as selection, crossover, and mutation \cite{poli_field_2008}. 
The output of the model is a symbolic regression formula that relates crab weight to its characteristics in a transparent and readable way.

\section{Explaining Sustainability and Healthiness}\label{sec:summary_sustainability}

A balanced use of data types and explanation methods is observed in sustainability and healthiness studies, with equal representation of pictorial and tabular data, along with one study utilizing time series data.
\Cref{tab:tbl_sustainability} summarizes the studies surveyed in the present section.


\begin{table}[htbp]
\centering
    \caption{Summary of the works introducing applying XAI for the Sustainability and Healthiness topic surveyed in \Cref{sec:summary_sustainability}, according to their data type and explanation type (labelled as ``Expl.\ type'').}
    \label{tab:tbl_sustainability}
\begin{tabularx}{\textwidth}{p{0.55\linewidth} p{0.2\linewidth} p{0.2\linewidth}}
    \textbf{Works} & \textbf{Data type} & \textbf{Expl.\ type} \\
    \midrule
    \cite{nr33,nr56,nr87,nr92,nr133,nr141,nr213} & \textcolor{pictorialdata}{Pictorial} & \textcolor{visualexpl}{Visual} \\

    \cite{nr101,nr110,nr155,nr162,nr167,nr180,nr191,nr193,nr226} & \textcolor{tabulardata}{Tabular} & \textcolor{numericalexpl}{Numerical} \\

    \cite{nr105,nr210} & \textcolor{timeseriesdata}{Time series} & \textcolor{numericalexpl}{Numerical} \\

    \cite{nr114} & \textcolor{tabulardata}{Tabular} & \textcolor{rulebasedexpl}{Rule-based} \\

    \cite{nr115} & \textcolor{timeseriesdata}{Time series} & \textcolor{rulebasedexpl}{Rule-based} \\

    \cite{nr134} & \textcolor{tabulardata}{Tabular} & \textcolor{mixedexpl}{Mixed} \\
\end{tabularx}
\end{table}

\subsection{Visual Explanation for the Sustainability and Healthiness topic}

\paragraph{Pictorial data.}
The works in this section used Grad-CAM as an XAI technique, confirming its widespread application in explaining solutions to Sustainability and Healthiness problems using pictorial data. 
\cite{nr33} used a CNN combined with a feature-based cascade classifier in pig face recognition. They employed Grad-CAM to verify that the model focuses on key facial features, offering a cost-effective alternative for animal identification in intensive farming. The paper contributes to the field of animal identification, improving welfare and non-invasive animal management practices.
\cite{nr92} utilized a CNN model, AlexNet, with UAV-based RGB imagery to predict forage biomass, with Grad-CAM confirming that the model accurately identified relevant regions for biomass prediction. \cite{nr87} proposed a CNN model to detect rice phenology stages using smartphone images. Grad-CAM showed that the model effectively recognized developmental stages, demonstrating the potential of using low-cost tools for real-time agricultural monitoring. 
\cite{nr56} introduced MSANet, a model combining multiscale attention and CNN layers for fruit recognition. Grad-CAM was used to interpret the model’s decisions, ensuring effective feature identification for robust fruit classification across applications. This work advances waste reduction through automated fruit detection, promoting environmental sustainability.
\cite{nr213} created a modular DL-based framework, called AMULET, that predicts plant traits from RGB images. The framework was tested on potato plants and on Arabidopsis thaliana, but could be extended to other plant species by predicting plant appearance. Two gradient-based XAI techniques, TorchGrad and Grad-CAM, were applied to explain the results.
Similarly, \cite{nr141} applied a Vision Transformer (ViT) model for plant seedling classification and used attention heatmaps to provide insights into the model’s decision-making process.
Lastly, \cite{nr133} developed a CNN-based system as an automated method for evaluating the precision of agricultural sprayers by detecting spray deposits, eliminating the need for manual tracers or water-sensitive papers. The study also employed an XAI pipeline---specifically Grad-CAM and Grad-CAM++---to interpret the CNN’s decision-making process, revealing key spatial filtering methods used for classification.

\subsection{Numerical Explanations}

\paragraph{Tabular data.}
The studies explored the application of ML and XAI techniques in health, food, and agriculture using tabular data. 
In this context, the works primarily address sustainable cultivation and livestock management practices, with particular attention to animal behavior.
\cite{nr191} integrated ML and DL models—including SVM, RF, and NNs—with LIME and SHAP to provide a transparent and efficient solution for crop yield prediction, focusing on automating agricultural processes and promoting sustainability.
\cite{nr110} combined genomic and environmental data to predict wheat yield using advanced DL frameworks.
DeepLift \citep{shrikumar_learning_2017} analysis revealed that environmental factors were more influential than genetics, highlighting the importance of integrating both data types for crop variety development.
\cite{nr162} proposed the use of a sensing agricultural robot that collects data such as temperature, humidity, and UV index to automatically forecast mulberry plant diseases by monitoring environmental conditions over time, leveraging LightGBM for prediction and SHAP for interpretability.
\cite{nr207} proposed a method to determine irrigation requirements for spinach using soil and climate data. The model was based on a stacked NN architecture, specifically a CNN and a Long Short-Term Memory network, to predict irrigation demands.
Similarly, \cite{nr180} introduced a real-time irrigation management system for paddy fields, utilizing a hybrid and ensemble feature extraction approach, called HyEn-X, combined with a Federated Learning-based framework, enabling decentralized learning for localized decision-making while preserving data privacy; SHAP was employed to enhance model interpretability.

\cite{nr167} exploited ML-based techniques integrated with LIME and SHAP to predict cattle behavior using sensor data collected from eighteen cows via accelerometers and pressure sensors, classifying behaviors into \emph{Other behavior}, \emph{Ruminating}, and \emph{Drinking/Eating}.
\cite{nr226} proposed a DL system that predicted the risk of ducks dying on arrival during transport in trucks. SHAP played an essential role in understanding the contribution of the features learned by the model.

\cite{nr101} focuses on human health and potential risks. The authors employed an RF model, using SHAP values to assess the impact of phenol-enriched olive oils on cardiometabolic health in hypercholesterolemic individuals. The study found that phenol-enriched oils significantly reduced serum metabolites associated with cardiovascular risk, indicating their potential as a treatment for cardiometabolic diseases.

A couple of studies focus on plant phenotypes. 
\cite{nr155} applied an RF model to predict almond shelling fraction using genotype data, with SHAP analysis providing insights into the genetic markers influencing shelling fraction and thereby supporting informed breeding strategies.
cite{nr193} developed a system for field-scale potato phenotyping by leveraging UAV-based multispectral imagery and machine learning. The authors engineered novel features from the spectral data and used XGBoost and RF for prediction. SHAP confirmed that the proposed approach provides a non-invasive and highly accurate method for estimating potato plant traits.

\paragraph{Time series data.}
Considering time series, \cite{nr105} proposed several ML models to predict individual pig growth trajectories from group-level weight data, reducing reliance on traditional, costly Radio Frequency Identification tracking. 
The RF model performed best, with an average Root Mean Square Error of \num{2.26} kg per pig. 
SHAP analysis highlighted weight and time differences as key predictors, supporting ML as a cost-effective alternative for growth estimation.
\cite{nr210} introduced a system for monitoring equine behavior based on wearable sensors and ML. In particular, they proposed using RF to predict animal behavior (e.g., walking, grazing, galloping, head shaking) using data collected from a neck-mounted accelerometer, gyroscope, and magnetometer. SHAP was used to explain the obtained results.

\subsection{Rule-based Explanation for the Sustainability and Healthiness topic}

\paragraph{Tabular data.}
Tabular data was explored in \cite{nr114}, where the authors proposed a system utilizing IoT data, encompassing crop types, soil characteristics, and weather conditions---to monitor the agricultural environment and alert farmers about necessary actions to maintain optimal crop conditions. This method, based on fuzzy logic and integrated with ML algorithms, detects anomalous data resulting from security breaches or hardware malfunctions.

Results indicated that the system effectively increased crop yields through real-time monitoring and decision-making based on IoT insights. The fuzzy logic framework enhanced system interpretability, making it user-friendly for farmers. Tested on maize, the system demonstrated high interpretability, accurate anomaly detection, and reliability in triggering appropriate actions.

\paragraph{Time series data.}
\cite{nr115} highlighted the need to monitor low-cost, automated, and interpretable irrigation systems using time series data. To address this, they proposed a new system called Vital, which integrates IoT sensors, a data management platform, and a fuzzy rule-based decision support system to automate irrigation. The system was evaluated through pilot cases and effectively automated the irrigation process, monitoring, and managing open-field installations that provided water.

\subsection{Mixed Explanation for the Sustainability and Healthiness topic}

\paragraph{Tabular data.}
In the field of Sustainability and Healthiness, only one study applies XAI techniques in conjunction with tabular data.
\cite{nr134} introduced AgriUXE, a digital platform that integrates XAI with multimodal data to enhance decision-making in smart farming, bridging the gap between AI-based agricultural solutions and farmers' understanding by providing tailored explanations based on IoT sensor data, remote sensing, and predictive models.
The authors presented an effective case study in viticulture by integrating various AI-based methods with multiple XAI techniques, including LIME and SHAP.

\section{Comparison and Insights}

\begin{figure}[!tp]
    \centering
    \includegraphics[width=\textwidth]{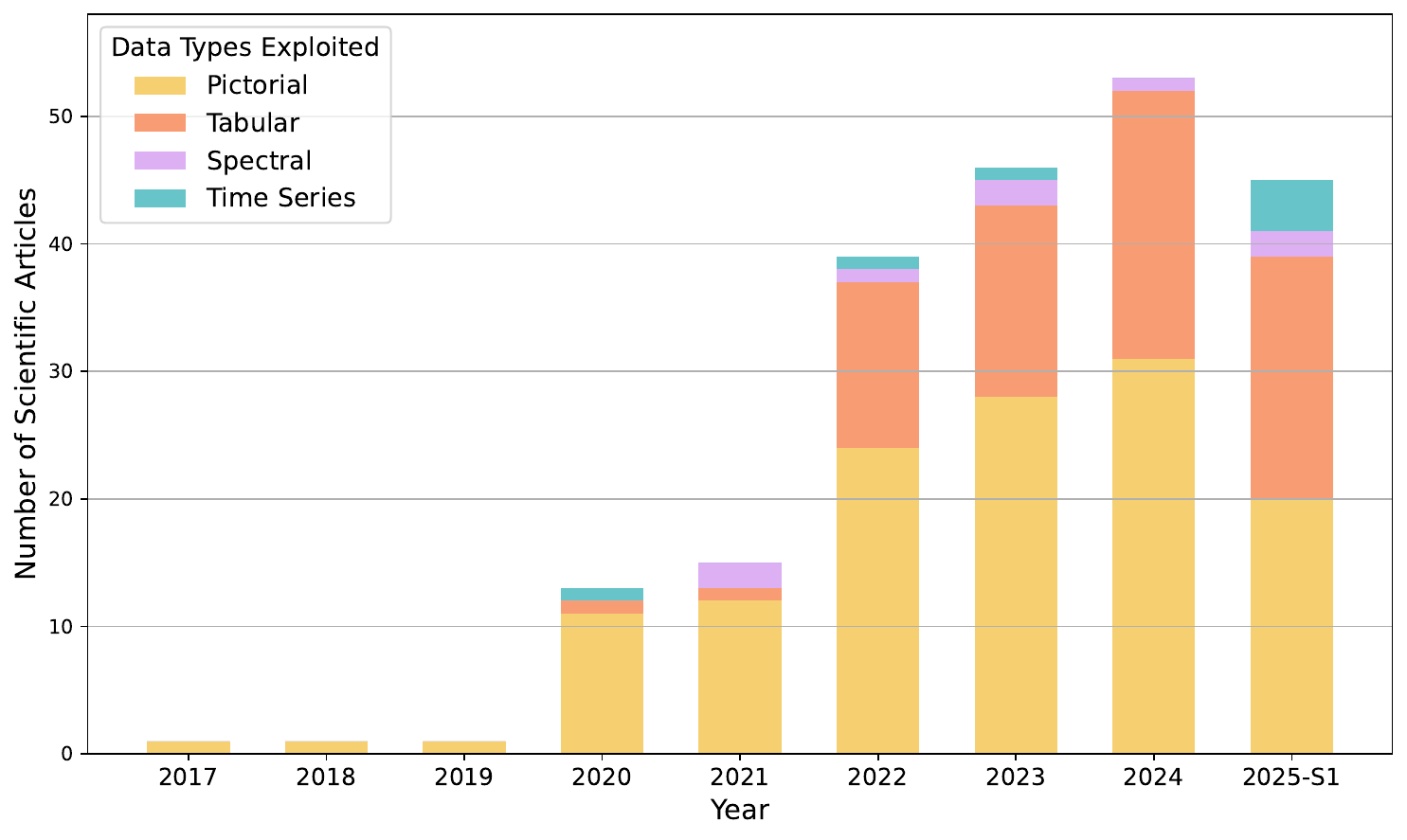}
    \caption{Distribution of surveyed papers by publication year and data type. The sharp increase from 2020 onwards reflects the growing interest in applying XAI techniques to food quality, with pictorial data dominating early research and tabular data gaining traction in recent years. Time series and spectral data remain underexplored despite their widespread use in physicochemical analysis.}
    \label{fig:years_exp}
\end{figure}

\Cref{fig:years_exp} showcases the number of scientific articles by year and data type.
Here, we can notice an increasing use of XAI in recent years, reflecting a significant rise in interest and application.
The data reveal a significant progression, reflecting an increasing awareness of the importance of transparency and interpretability in AI models within the food industry. 
Early research and the majority of studies have primarily focused on pictorial data, with growing attention to tabular data. 
It is important to note that time series and spectral data, which are widely used in physicochemical analysis, have not been extensively explored with XAI techniques.

In the field of food quality, significant opportunities for growth and development have recently emerged through AI as a powerful innovation tool, as highlighted by \cite{yi_recent_2024}. This work provides an overview of the various AI techniques available and applied to food quality, describing several notable studies that propose solutions to the challenges discussed in this review. 
Numerous studies in the literature focus on identifying and analyzing key applications of AI in food quality \citep{othman_artificial_2023}, as well as related areas like food processing \citep{jadhav_transformative_2025}, or trying to improve the entire food supply chain \citep{mavani_application_2022,manning_artificial_2022}. 
Others explore specific techniques, such as computer vision \citep{kakani_critical_2020}, which is popular for handling pictorial data types.
Through this analysis, we can observe a steady increase in the use of AI in food engineering, with a growing number of innovations being tested and introduced. 
This trend reflects the rising popularity of AI and the continuous improvements in the versatility and accuracy of the models.
However, as the \Cref{fig:years_exp} shows, adopting XAI techniques has not grown at the same pace. 
Only a small fraction of studies that employ AI also integrate XAI methods. 
This can be attributed to researchers' focus on developing highly efficient and accurate models to solve the proposed problems. 
Current food research aims to identify new applications and refine existing models to enhance accuracy. 
The need for model interpretability becomes less urgent once satisfactory performance levels are achieved. 

\begin{figure}[!t]
    \centering
    \includegraphics[width=\textwidth]{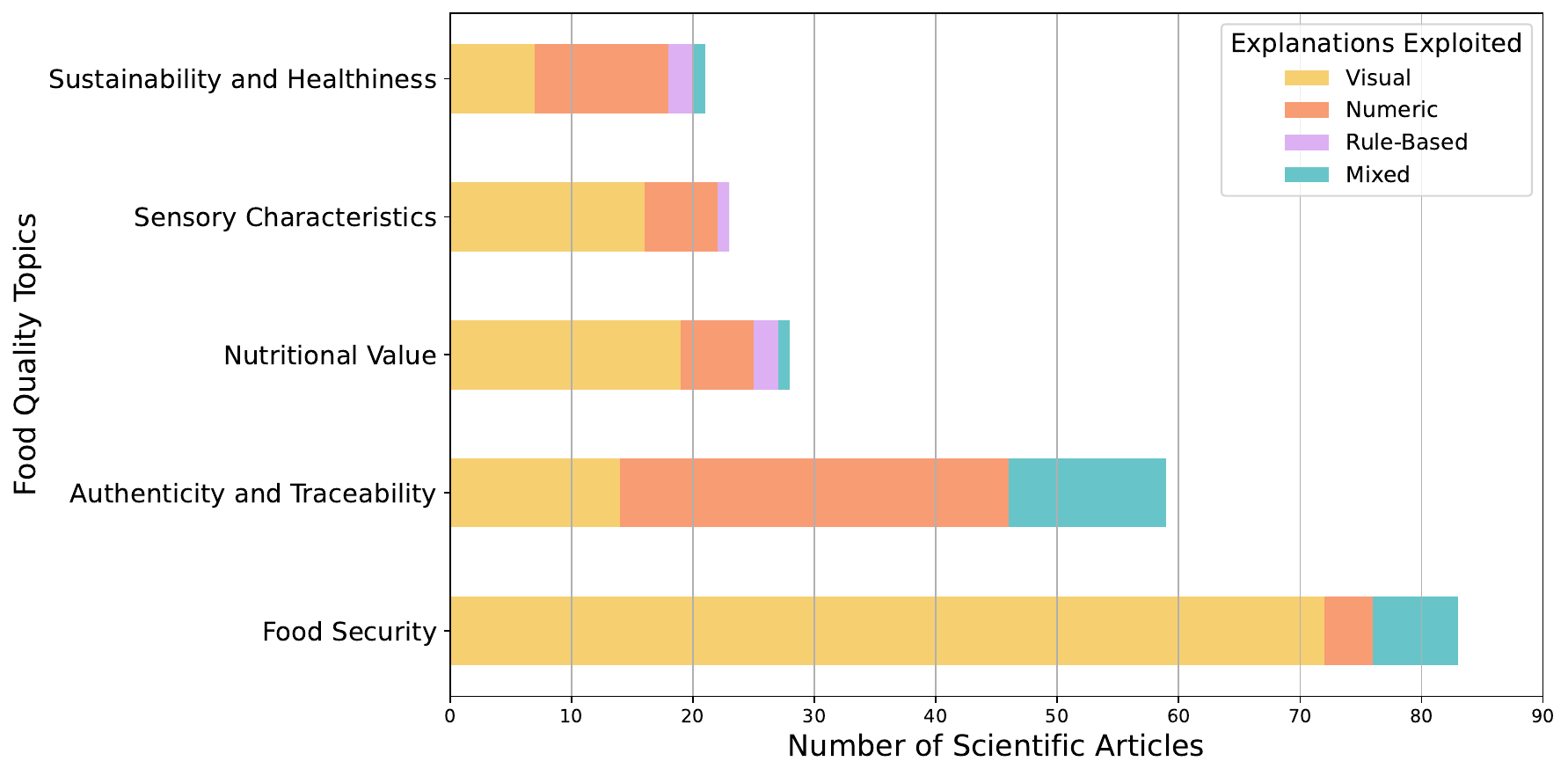}
    \caption{Distribution of surveyed papers by food quality topic and explanation type. Food Security emerges as the dominant application area, followed by Authenticity and Traceability and Nutritional Value. Sensory Characteristics and Sustainability and Healthiness remain underexplored, and a consistent gap is observed in the use of XAI for spectral and time series data across all topics.}
    \label{fig:paps}
\end{figure}

The examination of articles, as shown in \Cref{fig:paps}, reveals that Food Security is the most prominent topic of XAI application. This theme is central to the majority of studies reviewed, closely followed by Authenticity and Traceability and Nutritional Value, both of which are also important in this research domain. In contrast, topics such as Sensory Characteristics and Sustainability and Healthiness are less frequently explored, indicating a lower level of interest from the scientific community in applying XAI techniques to these areas. Again, we observed a gap in the interpretation of spectral and time series data using XAI methods.

\begin{figure}[!t]
    \centering
    \includegraphics[width=1\textwidth]{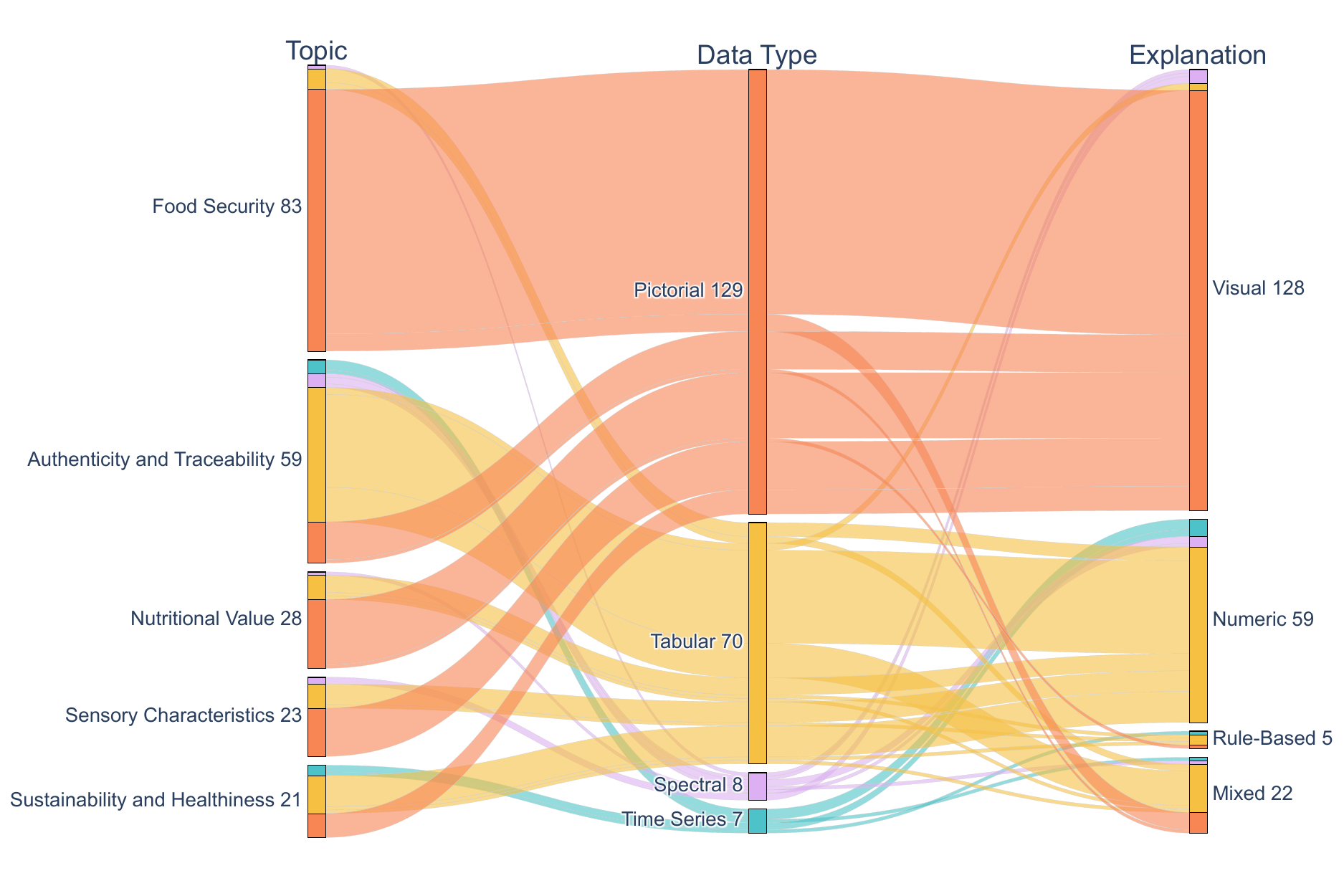}
    \caption{Distribution of works surveyed. While the distribution across topics is rather uniform, most of the works we survey concentrate on pictorial data and visual explanations, while a smaller portion of research deals with tabular data and numerical explanations.}
    \label{fig:alluvial}
\end{figure}

The papers expose that pictorial data are the most frequently used data type, followed closely by tabular data, which is also widely utilized, as shown in \Cref{fig:alluvial}. In contrast, spectral and time series data are utilized much less frequently. 
The prevalence of pictorial data can be attributed to several factors, which can be detected by observing the second half of the plot. 
XAI techniques that provide visual explanations, such as CAM-based methods, are widely employed in the literature, as noted by \cite{vilone_classification_2021}. 
These techniques are highly popular because they provide readily interpretable visual explanations, often as heatmaps, making them ideal for users with limited experience in the analyzed data who still need an intuitive, immediate understanding of the decision-making processes of the image analysis model.
This utility justifies why a significant portion of the surveyed papers rely on them, consequently requiring pictorial data.
In contrast, techniques that offer numerical explanations, though popular, are not as easily interpretable and are therefore primarily used for analyzing tabular data. 
Rule-based explanation techniques, on the other hand, are less common and thus less frequently exploited.

\begin{figure}[t!]
  \centering
  \subfloat[]{\includegraphics[width=0.45\textwidth]{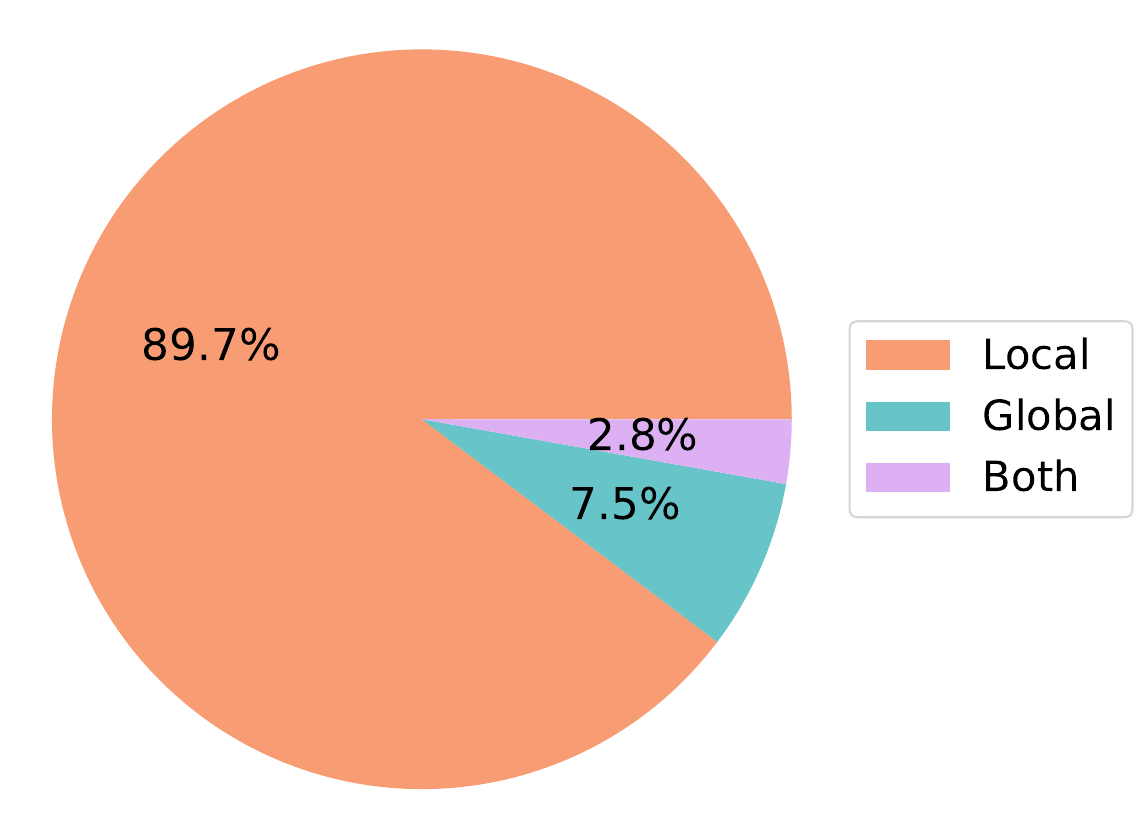}}\hfill
  \subfloat[]{\includegraphics[width=0.52\textwidth]{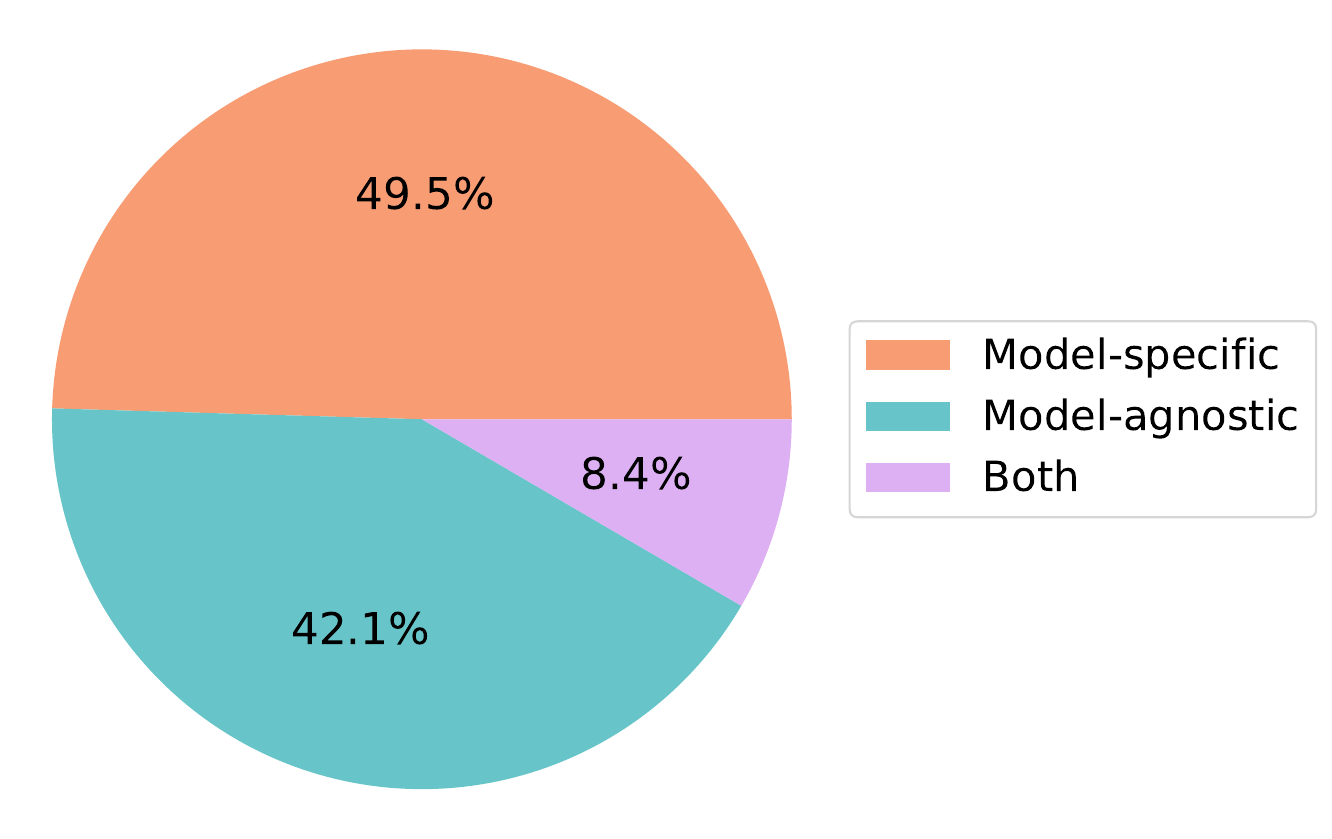}}\hfill
  \caption{Distribution of surveyed papers that utilize global XAI techniques versus local XAI techniques (a), and the percentages of papers that apply model-agnostic XAI techniques compared to model-specific XAI techniques (b). In both charts, the Both sector indicates that there are works employing multiple techniques of different types. Observing the first graph, it becomes evident that most of the studies reviewed employ local XAI techniques, such as the widely recognized SHAP and Grad-CAM. In contrast, global techniques appear to be less commonly used. The second graph confirms this observation, since SHAP is model-agnostic, while Grad-CAM is model-specific, highlighting an even distribution of the two most popular techniques across the various works.}
  \label{fig:pie_charts}
\end{figure}

\Cref{fig:pie_charts}a highlights a clear preference for local methods over global ones. 
This preference is driven by the popularity of techniques like LIME, SHAP, and Grad-CAM in the reviewed works, all of which are local methods. These methods are simple to apply, offer easily interpretable explanations, and are particularly useful for understanding the model’s decision-making in individual cases.
In contrast, global methods, which are more suited for gaining an overall view of the model’s decision-making process, are less frequently used due to their complexity, especially when applied to highly intricate models. 

The prominence of model-specific techniques, as shown in the \Cref{fig:pie_charts}b, is largely due to the widespread use of CAM-based methods.
The analysis of the papers suggests that CNNs are the most commonly used approach for pictorial data, while CAM-based techniques are the most straightforward choice for explaining these models.
In contrast, LIME and SHAP, the other two most commonly used methods, are model-agnostic. 
In this case, there is no clear preference between the two types of XAI; instead, the focus tends to be on certain specific techniques.

Lastly, it is worth noting that few papers employ more than one XAI technique, which limits the understanding of the model's decision-making process to a partial view. 
This is especially true when using local techniques, which provide explanations for individual samples without offering insight into the model’s broader decision-making patterns.

\subsection{Critical evaluation of XAI methods}\label{sec:critical_eval_xai}

While XAI techniques can enhance transparency and support model validation, their outputs should not be interpreted as ground truth.
In food engineering, where decisions may affect safety, compliance, and product quality, the reliability of explanations becomes a central concern.
In the following section, we examine the principal limitations and practical implications of the most widely adopted explainability techniques, as identified through the systematic analysis of the surveyed literature. These limitations should therefore not be regarded as inherent properties of the techniques themselves or as universally valid claims, but rather as recurring patterns that emerged from the examination of the reviewed works.

\paragraph{Reliability and stability of model-agnostic explainers (LIME/SHAP).}
Model-agnostic explainers such as LIME and SHAP are widely adopted because they can be applied to different black-box models and data types.
However, both methods are susceptible to \emph{instability} and \emph{sensitivity} with respect to design choices \cite{nauta_anecdotal_2023}.
For LIME, explanations depend on how the local neighborhood is sampled, including the number of perturbed instances, the kernel width, and the distance metric, as well as on the choice of perturbation operator. Small variations in these settings can yield different surrogate models and, consequently, divergent feature attributions. This issue is particularly problematic in high-dimensional settings such as spectral vectors or image inputs, where perturbations can generate unrealistic samples that lie outside the data manifold, thereby undermining the quality of the local surrogate.

Similarly, SHAP explanations are sensitive to the background or reference distribution used to simulate feature removal, as well as to the assumptions made regarding feature dependence. In food datasets, predictors are frequently correlated, as is the case with physicochemical measurements, spectral wavelengths, and environmental covariates. Independence-based approximations can therefore distribute attribution credit across correlated features in a manner that does not accurately reflect the model's actual decision logic. Furthermore, SHAP values may shift when the baseline is altered, for instance, when substituting the training-set mean with a task-specific reference point, which can undermine interpretability for domain practitioners.

When employing LIME or SHAP in food quality applications, it is therefore important to explicitly report the perturbation strategy and baseline selection, and to treat the resulting attributions as hypotheses to be validated rather than as definitive explanations.

\paragraph{Noise and sensitivity in saliency-based visual explanations.}
Saliency and CAM-based techniques, including CAM, Grad-CAM, and their variants, are widely used for visual data because they produce heatmaps that are intuitive for human stakeholders. Nevertheless, saliency maps can be noisy and sensitive to model parameters, input preprocessing choices, and small input perturbations.

In practice, heatmaps may highlight broad or diffuse regions and spurious patterns such as backgrounds, lighting artifacts, packaging elements, or imaging device signatures, particularly when models exploit shortcut cues during training. This risk is especially relevant in food inspection scenarios where acquisition conditions vary across illumination, moisture levels, occlusion, and camera angle, and where visually salient regions do not necessarily correspond to causally meaningful factors underlying the quality attribute of interest. Additionally, CAM-based methods typically provide coarse spatial localization and should not be interpreted as precise segmentation of defects or contaminated areas.

As a mitigation strategy, studies employing visual explanation methods can complement heatmap analysis with sanity checks such as randomization tests, sensitivity analysis under mild input perturbations, and systematic comparisons across multiple explainers to assess the consistency and reliability of the resulting attributions.

\paragraph{Recommendations and limitations for the effective application of XAI techniques.}
The choice of XAI method should be aligned with the data modality, the model architecture being employed, and the informational needs of the relevant stakeholders. For instance, the present survey reveals that when NNs are applied to pictorial data, the selection of an XAI technique falls almost exclusively on Grad-CAM, despite the fact that other established methods perform comparably well in this setting. A more informed use of the available XAI techniques, grounded in a clear understanding of what each method actually computes, can help overcome the substantial difficulties that commonly limit their effective adoption.

One such difficulty is the gap between a plausible explanation, that is, one that appears convincing to a human observer, and an explanation that genuinely reflects the model's true decision process. In some cases, an explanation may seem counterintuitive or unconvincing while nonetheless accurately capturing the model's learned behavior. In others, an explanation may appear well-aligned with user expectations while failing to correspond to what the model has actually learned \citep{arrighi_explainable_2023}. In food engineering, this gap is amplified by domain complexity and the practical urgency of operational decisions.

We therefore recommend treating XAI as a diagnostic layer rather than a source of ground-truth justifications. Explanations should be triangulated against domain knowledge, controlled experiments, and independent measurements where feasible. Furthermore, combining explanations from multiple methodological families, such as SHAP and PDP for tabular data, or Grad-CAM paired with a perturbation-based method for image data, provides a more robust characterization of model behavior and reduces the risk of over-interpreting artifacts produced by any single XAI approach.

Finally, we recommend extending the application of XAI techniques to every stage of the pipeline in which an ML or AI model is employed. In practice, preprocessing and data cleaning steps frequently rely on models that themselves operate as black boxes, yet their outputs propagate through the entire workflow and can substantially influence the final result. We therefore advocate for the consistent application of explainability methods to these upstream stages as well, so as to ensure transparency and reliability throughout the full analytical pipeline rather than solely at the prediction stage \cite{ceschin_extending_2026}.

\paragraph{Recommendations for reporting and validation.}
To improve transparency and reproducibility, we recommend that food quality studies employing explainability methods explicitly report the explainer configuration, including the perturbation operator, kernel width, and baseline choices, sensitivity analyses assessing the stability of explanations under resampling or mild input perturbations, and quantitative metrics that justify both the appropriateness of the chosen technique and the validity of the resulting explanations, such as faithfulness-oriented evaluations through removal-based assessments and sanity checks for saliency maps, along the lines of the evaluation framework proposed by \cite{nauta_anecdotal_2023}.

\section{Open Challenges and Future Directions}

Based on the analysis presented in this survey, two key directions emerge for improving the food industry's supply chain through the integration of AI.
The first direction arises from the growing prevalence of AI technologies in the world and the resulting adaptation by both the industry and society. This is a trend that the food industry should also embrace.
The second direction focuses on the development of new technologies in the field of XAI. These technologies have the potential to enhance the application of AI within the food industry and expand its possible uses.
In the following subsections, we will discuss these two directions in more detail.

To provide a concise roadmap, we summarize the five principal research gaps emerging from the surveyed literature.
\begin{itemize}[leftmargin=25pt, itemindent=0pt, labelsep=0.5em, topsep=0pt, partopsep=0pt]
\item[(G1)] Real-time interpretable imaging at scale.
\item[(G2)] Actionable counterfactual explanations for process and formulation decisions.
\item[(G3)] Human-in-the-loop deployment and validation protocols.
\item[(G4)] Benchmarks and formal evaluation metrics for explanation quality.
\item[(G5)] Regulatory compliance, traceability, and post-deployment monitoring for high-stakes applications.
\end{itemize}
These gaps are revisited throughout the following subsections and collectively define a prioritized set of directions for both future research and industrial adoption.

\subsection{Integrating AI into the Food Engineering}

Despite the increasing interest in interpretable modeling, the adoption of XAI within food engineering remains relatively low. As discussed by \cite{manning_artificial_2022} and evidenced throughout the surveyed literature, this limited uptake can be attributed to several interrelated factors.

Some limitations are primarily practical. 
First, the effective application of XAI methods requires multidisciplinary expertise, combining data science proficiency or deep domain knowledge in food technology, which is still scarce across research groups and industrial teams. 
Moreover, the computational and implementation costs associated with advanced interpretability frameworks---such as SHAP, Grad-CAM, or hybrid visualization-based pipelines---remain prohibitive for many laboratories and small-scale producers, particularly when dealing with high-dimensional spectral or multimodal datasets.
Additionally, there is an overall \emph{lack of formal evaluation of explanations}, a topic that poses several challenges, mainly due to the missing---or difficult to obtain---ground truths.
However, as noted by \cite{nauta_anecdotal_2023}, it is paramount that the quality of explanations be evaluated in a formal way, and not only in an anecdotal manner or via user studies.

From a deployment standpoint, a critical bottleneck is represented by (G1) real-time interpretable imaging at scale. Many food industry scenarios rely on high-throughput vision systems for tasks such as defect detection, contamination screening, and product grading, where explanations must be generated under strict latency constraints and remain stable across varying acquisition conditions, including illumination, camera geometry, and packaging reflections, as well as across seasonal and batch-to-batch changes. Achieving low-latency and robust explanations for imaging models remains an open engineering and research challenge, particularly when transitioning from controlled benchmark datasets to continuously evolving production environments.

The regulatory landscape of the food sector also imposes barriers to the use of AI. Strict safety standards and validation requirements often discourage the deployment of novel AI-driven decision systems unless their traceability and reliability can be fully demonstrated \citep{manning_artificial_2022,da2024deep,demoraes2024interpretation}.
On the side of AI-specific regulations, the EU AI Act \cite{act2024eu} mandates \emph{transparency} as a fundamental property of \emph{High-risk} AI systems.
Specifically, in the food sector, \cite{val2025eu} argues that many AI applications could fall under the High-risk umbrella due to their direct impact on human health or other societal impacts.
While it could be argued that transparency is not necessarily solely connected with explainability \cite{panigutti2023role}, XAI could still be used as a tool to enhance human oversight of these systems.
The extent to which XAI provides a quantifiable contribution to transparency, as mandated by the EU AI Act, is to be determined by \emph{conformity assessments}.
The regulation mandates that these be carried out, possibly by third-party entities, to determine whether these systems comply with the Act.
Thus, any consideration of guidelines for XAI in food engineering will have to be determined on a case-by-case basis.

This directly connects to (G5) regulatory compliance, traceability, and auditing. When artificial intelligence systems influence decisions pertaining to food safety, product labeling, or fraud detection, stakeholders require traceable evidence linking model outputs to documented decision criteria, in a form that can support inspection, reporting, and conformity assessments. As a consequence, XAI should not be treated merely as a post-hoc visualization layer, but rather as an integral component of a broader compliance workflow that encompasses documentation, model versioning, and repeatable explanation generation under controlled and standardized conditions.

Remaining in the context of AI regulations, the US Food and Drug Administration (FDA) also mandates requirements for the usage of AI systems, albeit for the development of ``\emph{drugs and biological products}'' \cite{fda2025considerationsAI}.
Similar to the EU AI Act, these considerations propose a framework based on risk and context of use to determine the level of transparency of the system.
This could translate seamlessly to food applications, given, as previously stated, the potential for high risk in some of the applications surveyed in the present work.
In addition to this risk-based framework, the FDA also emphasizes the need to monitor the AI system \emph{after} deployment.
XAI techniques, in this context, have been shown to help identify post-deployment phenomena such as distribution shifts \cite{apicella2022xai} and concept drift \cite{lee2023explainable,teixeira2025detecting}.

Post-deployment monitoring is also tightly coupled with (G3) human-in-the-loop deployment and validation. In real production environments, explanations must be reviewed and acted upon by inspectors and quality managers, and the system should support structured feedback loops encompassing the flagging of systematic explanation drift, the triggering of re-calibration procedures, and the activation of escalation protocols when anomalous behavior is detected. Designing reliable operational workflows that integrate automated alarms with human verification, therefore, remains a central requirement for the reliable adoption of XAI in industrial food quality control.

In conclusion, while AI has become an increasingly essential tool throughout the food supply chain, its use remains intrinsically limited by its opaque nature. 
XAI, which offers an effective response to the black-box problem by providing clarity and transparency, also presents significant challenges. 
Due to this complexity, research in the field of XAI is highly active and rapidly evolving. Existing techniques are continually being updated and improved, while new methods are consistently emerging. Researchers should stay informed about these advancements, as applying and experimentally evaluating new techniques can help refine them and encourage their adoption.

\subsection{Advancing XAI Technologies for the Food Supply Chain}

Through the comparative analysis of the studies surveyed and an evaluation of their statistical insights, we have identified several opportunities for further research and future XAI applications in the field of food quality.

By studying the work of \cite{bodria_benchmarking_2023}, we can observe that, to our knowledge, several types of XAI techniques have not been utilized in the analyzed works, yet they could prove extremely useful in various contexts.
For instance, rule-based techniques, which we appreciated in just some studies \cite{nr1,nr114,nr115,nr132,nr209}, offer the potential to uncover causal relationships between the physicochemical properties of food products, adding depth to the interpretation of these behaviours and aiding in the development of more interpretable models. 

We can also observe that, according to \cite{nr17} and to the works summarized by \cite{longo2024explainable}, the explanations provided by many XAI methods are often difficult to interpret and require the expertise of a field specialist, turning the explanations themself into additional steps to be deciphered. 
To address this, recent advancements in XAI research propose new solutions, such as the development of frameworks, metrics to evaluate model outputs \citep{arras_clevr-xai_2022, nauta_anecdotal_2023}, and the use of \emph{generative AI} to simplify and clarify the explanations provided by existing methods, e.g., by using Large Language Models to provide text-based explanations that helps interpreting explanations \cite{silvestri2025survey}.
Moreover, utilizing various methods helps achieve a thorough understanding of a model's decision-making process.
Specifically, by adopting a \emph{glocal} approach \citep{longo2024explainable}, which combines both local and global explanations of the same model, more complete and user-friendly explanations can be achieved.

Among the various new techniques for which no applications were identified in this survey, \emph{concept-based learning}, algorithms represent a popular category of methods that can be used to explain model predictions in terms of adjectives, concepts, or abstractions easily understood by humans \citep{kim_interpretability_2017}.

Additionally, \emph{counterfactual explanations}, extensively studied in XAI research \cite{longo2024explainable}, can significantly support \emph{in silico} food simulation research. 
These techniques help to understand how variations in ingredients, environmental conditions, and processes impact food quality, taste, or nutritional profile.
Counterfactual explanations also allow researchers to examine how specific ingredient changes might influence shelf life. 
By simulating alternative pathways without requiring costly or time-intensive experiments, counterfactual methods can guide decision-making, optimize formulations, and enhance the accuracy of outcome predictions.

This motivates the need for (G2) actionable counterfactual explanations. Beyond answering why a particular decision was reached, many food engineering use cases require guidance on what should change in order to achieve a desired outcome, such as improving a quality grade, extending shelf life, or reducing defect probability. Future work should therefore focus on counterfactual generation that is (i) feasible under process constraints, including bounded temperature ranges, permitted additives, and production tolerances, (ii) consistent with the interdependencies among correlated variables such as moisture and texture interactions, and (iii) validated against domain knowledge and, where possible, confirmatory experimental evidence.

Another potential area for improvement is the use of \emph{hybrid systems} that combine black-box models with white-box components, as seen in Neurosymbolic AI \cite{bhuyan2024neuro}.
Another strategy could involve training black-box models on engineered features rather than on raw data, a common approach in time-series analysis \cite{kim2025comprehensive}. This method may enhance interpretability for experts in the field.

An alternative emerging approach in food quality assessments is data fusion, as it integrates multiple types of data, such as chemical, physical, and sensory information, to make more comprehensive decisions about food products. 
This fusion of diverse data types enables a richer analysis but makes understanding the outcomes more challenging. 
To address this complexity, \emph{hierarchical-based} explainability approaches could be proposed to break down the contribution of each data type to the final decision.
Although not yet well-defined, this type of XAI could offer a viable solution for explaining models that integrate multiple data types. By introducing a hierarchy within explanations, it becomes possible to discern the contribution of each data type to the overall result, clarifying how the combined dataset influences the model’s decisions.

In summary, these approaches provide ways to enhance the reliability and transparency of AI algorithms across different areas of the food industry. 
As highlighted by \cite{miller2019explanation}, XAI should be implemented in human-in-the-loop systems primarily as diagnostic tools. This integration can help improve human reliability by allowing users to identify mistakes or biases in the relationships learned by these systems.
However, these tools offer limited support for real-time applications aimed at increasing efficiency. One potential application for XAI in real-time systems would be to randomly sample predictions from the model, generate explanations for those predictions, and have experts evaluate them. This process could help uncover prediction errors and prompt necessary interventions within the system.

This direction is closely aligned with (G3) human-in-the-loop deployment and (G4) benchmarks and formal evaluation metrics. Robust adoption requires both operational protocols and measurable quality criteria for assessing the validity of explanations. Beyond anecdotal examples, the research community needs shared benchmarks, standardized reporting of explainer configurations including baseline choices and perturbation strategies, and evaluation suites capturing key properties such as faithfulness, stability, and human usability. In food quality applications specifically, benchmark design should explicitly incorporate batch effects, seasonal variability, and acquisition shifts arising from sensor and imaging condition changes, to reflect realistic industrial distributions rather than static and controlled laboratory settings.

Finally, the expanding range of XAI methods offers promising opportunities to analyze underexplored data types. 
Spectral data, commonly used in physicochemical analysis, has yet to be adequately addressed by current XAI techniques.
Closing this gap will require the development of tailored approaches to interpret this complex data type. 
Additionally, there are relatively few XAI methods specifically designed to explain AI models used in time series analysis, as confirmed by \cite{bodria_benchmarking_2023}. 

Overall, addressing (G1) through (G5) jointly is essential for advancing from explanatory prototypes toward reliable and deployable systems. Real-time explanation generation must be paired with actionable counterfactual guidance, human-centered validation loops, benchmark-driven evaluation, and compliance-ready documentation and monitoring frameworks. Progress on any single gap in isolation is unlikely to be sufficient; it is their coordinated resolution that will enable the reliable integration of XAI into operational food quality control pipelines.

\section{Conclusion}
EXplainable Artificial Intelligence (XAI) techniques have emerged as important tools for enhancing the transparency and auditability of AI models, supporting the production of reliable and understandable outcomes.
These requirements are essential in food engineering, as food is a fundamental aspect of human life, and its quality and safety need to be studied with careful attention.
In this survey, we aimed to bridge the gap between these two disciplines by emphasizing the importance of XAI techniques and offering practical insights into both domains.

This work targets researchers and professionals in food science and technology interested in using XAI to enhance model transparency and regulatory compliance in food applications. 
It also serves as a reference for AI developers and policymakers focused on creating explainable systems in quality control.

Our comprehensive review is, to our knowledge, the first to connect the fields of XAI and food by examining how explainability techniques can be applied to food quality.
We analyzed a more than 200 studies from the literature and categorized them according to the types of data utilized---tabular, pictorial, spectral, and time series---and the forms of explainability provided, including numerical, rule-based, visual, and mixed explanations.
Additionally, we proposed a food quality taxonomy to contextualize the research, focusing on key areas such as food safety, nutritional value, sensory attributes, authenticity and traceability, and sustainability and healthiness. This taxonomy aims to unify existing literature and provide a consistent overview of the current state of XAI applications in food quality analysis.

Finally, we conducted a comparison of the studies to uncover valuable insights, identifying main trends, strengths, and divergences in the current research landscape. 
This analysis enabled us to map the current use of XAI in food quality applications, highlighting both its well-established applications and the areas that require further exploration. In particular, two key directions emerge: the broader adoption of AI (and XAI) across the food supply chain, and the development of more advanced XAI methods to enhance the transparency and impact of these applications.


\section*{Statements and Declarations}

The authors declare that they have no known competing financial interests or personal relationships that could have appeared to influence the work reported in this paper.

\section*{Declaration of generative AI and AI-assisted technologies in the writing process}

During the preparation of this work, the authors used ChatGPT-4o in order to improve readability. After using this tool/service, the authors reviewed and edited the content as needed and take full responsibility for the content of the published article.

\section*{Acknowledgments}

This work was supported by ASAC s.r.l., which funded the research fellowship of Leonardo Arrighi. The authors gratefully acknowledge this support. This study was also partially funded by the Coordination for the Improvement of Higher Education Personnel – Brazil (CAPES) – Finance Code 001; the São Paulo Research Foundation (FAPESP), under project numbers 2019/27354-3, 2019/03812-2, and 2023/07385-7; and the National Council for Scientific and Technological Development – Brazil (CNPq), under project numbers 140914/2021-8 and 307094/2021-9.

\bibliographystyle{unsrtnat}
\bibliography{references}

\end{document}